\documentclass{article}

\usepackage[preprint]{neurips_2024}

\usepackage[utf8]{inputenc} 
\usepackage[T1]{fontenc}    
\usepackage{hyperref}       
\usepackage{url}            
\usepackage{booktabs}       
\usepackage{amsfonts}       
\usepackage{nicefrac}       
\usepackage{microtype}      
\usepackage{amsmath}
\usepackage{cleveref}
\usepackage{eccvabbrv}
\usepackage{graphicx}
\usepackage[table]{xcolor}
\usepackage{subcaption}

\definecolor{verylightgray}{gray}{0.9}

\title{PerCo (SD): Open Perceptual Compression}

\author{%
  \textbf{Nikolai Körber}$^{1, 2}$ \quad
  \textbf{Eduard Kromer}$^{2}$ \quad
  \textbf{Andreas Siebert}$^{2}$ \quad \\
  \textbf{Sascha Hauke}$^{2}$ \quad
  \textbf{Daniel Mueller-Gritschneder}$^{3}$ \quad
  \textbf{Björn Schuller}$^{1}$ \\
  \\
  \textsuperscript{1}Technical University of Munich, Munich, Germany \\
  \textsuperscript{2}University of Applied Sciences Landshut, Landshut, Germany \\
  \textsuperscript{3}TU Wien, Vienna, Austria \\
}

\begin{document}

\maketitle

\begin{abstract}

  We introduce PerCo (SD), a perceptual image compression method based on Stable Diffusion v2.1, targeting the ultra-low bit range. PerCo (SD) serves as an open and competitive alternative to the state-of-the-art method PerCo, which relies on a proprietary variant of GLIDE and remains closed to the public. In this work, we review the theoretical foundations, discuss key engineering decisions in adapting PerCo to the Stable Diffusion ecosystem, and provide a comprehensive comparison, both quantitatively and qualitatively. On the MSCOCO-30k dataset, PerCo (SD) demonstrates improved perceptual characteristics at the cost of higher distortion. We partly attribute this gap to the different model capacities being used ($866$M vs. $1.4$B). We hope our work contributes to a deeper understanding of the underlying mechanisms and paves the way for future advancements in the field. Code and trained models will be released at \url{https://github.com/Nikolai10/PerCo}. 
  
\end{abstract}

\section{Introduction}

Perceptual compression, sometimes referred to as generative compression~\cite{Agustsson_2019_ICCV, mentzer2020high} or distribution-preserving compression~\cite{NEURIPS2018_801fd8c2}, refers to a class of neural image compression techniques that incorporate generative models (\eg, generative adversarial networks~\cite{NIPS2014_5ca3e9b1}, diffusion models~\cite{pmlr-v37-sohl-dickstein15, NEURIPS2020_4c5bcfec}) into their learning objective. Unlike traditional codecs such as JPEG, they additionally constrain the reconstructions to follow their underlying data distribution~\cite{ICML-2019-BlauM}. By leveraging powerful generative priors, missing details, such as textures, can be realistically synthesized, thus achieving higher perceptual quality at even lower bit rates. These characteristics make these methods particularly appealing for storage- and bandwidth-constrained applications.

Recently, foundation models~\cite{Bommasani2021FoundationModels_short}, large-scale machine learning models trained on broad data at scale, have shown great potential in their adaption to a wide variety of downstream tasks, including ultra-low bit-rate perceptual image compression~\cite{pan_2022, lei2023text+sketch, careil2024towards}. Notably, PerCo~\cite{careil2024towards}, the current state-of-the-art, is the first method to explore bit-rates from $0.1$ down to $0.003$bpp. For example, a bit-rate of $0.003$bpp translates to approximately $115$ bytes for an image of VGA resolution ($480\times640$), which is less the size of a tweet. This is essentially achieved by extending the conditioning mechanism of a pre-trained text-conditional latent diffusion model (LDM) with vector-quantized hyper-latent features. In other words, only a short text description and a compressed image representation are required for decoding. Despite its great potential and fascinating results, PerCo has not been made publicly available. This is arguably due to the fact that PerCo relies on a proprietary LDM based on GLIDE~\cite{pmlr-v162-nichol22a}. 

\begin{figure*}[tb]
    \setlength{\tabcolsep}{1.0pt}  
    \renewcommand{\arraystretch}{1.0}  
    \centering
    \scriptsize
    \begin{tabular}{cccc|c}
        \toprule
        Original & PICS ($0.0281$bpp) & VTM-20.0 ($0.025$bpp) & PerCo ($0.0032$bpp) & Ours ($0.0031$bpp) \\
        \midrule

        \begin{subfigure}{0.195\textwidth}
            \centering
            \includegraphics[width=\linewidth]{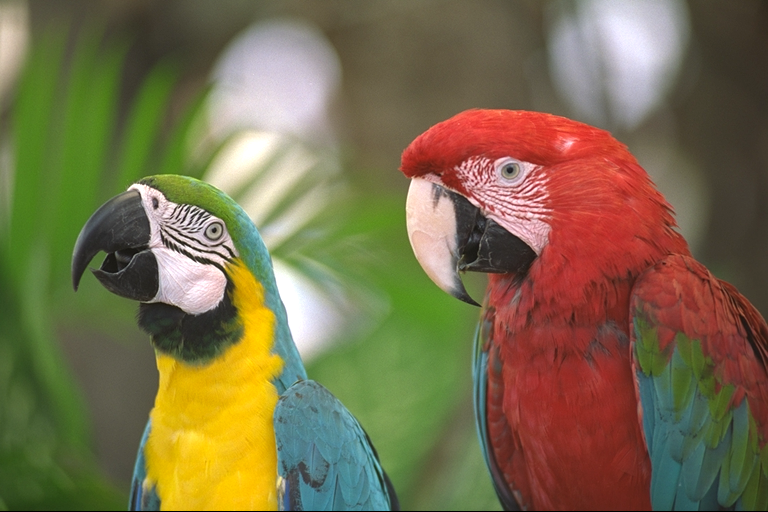}
        \end{subfigure}
        &
        \begin{subfigure}{0.195\textwidth}
            \centering
            \includegraphics[width=\linewidth]{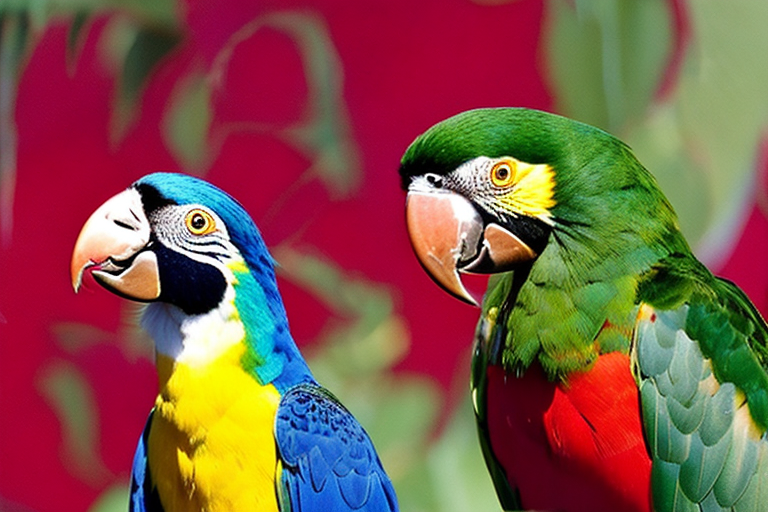}
        \end{subfigure}
        &
        \begin{subfigure}{0.195\textwidth}
            \centering
            \includegraphics[width=\linewidth]{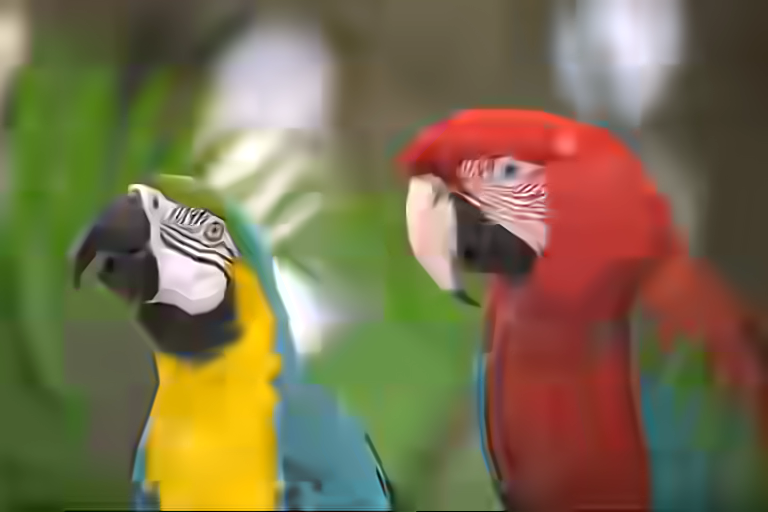}
        \end{subfigure}
        &
        \begin{subfigure}{0.195\textwidth}
            \centering
            \includegraphics[width=\linewidth]{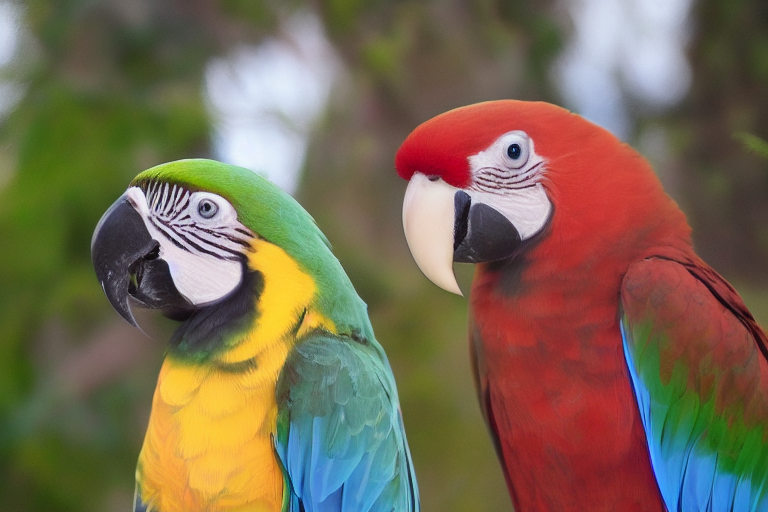}
        \end{subfigure}
        &
        \begin{subfigure}{0.195\textwidth}
            \centering
            \includegraphics[width=\linewidth]{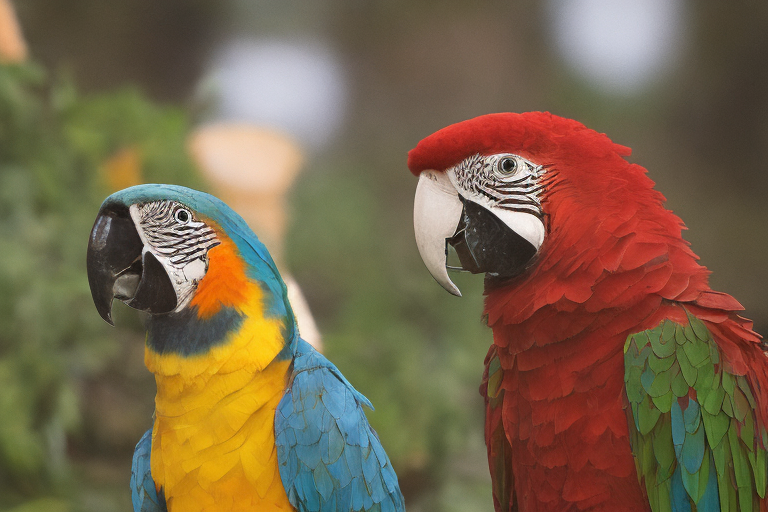}
        \end{subfigure} \\
        
        \begin{subfigure}{0.195\textwidth}
            \centering
            \includegraphics[width=\linewidth]{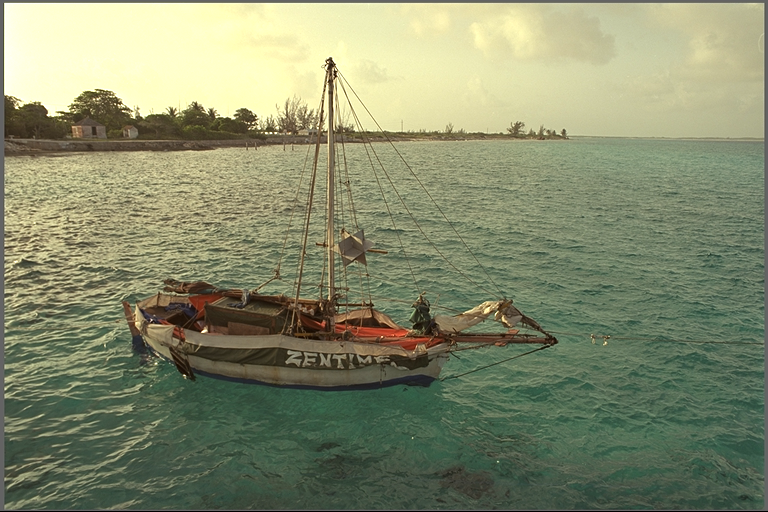}
        \end{subfigure}
        &
        \begin{subfigure}{0.195\textwidth}
            \centering
            \includegraphics[width=\linewidth]{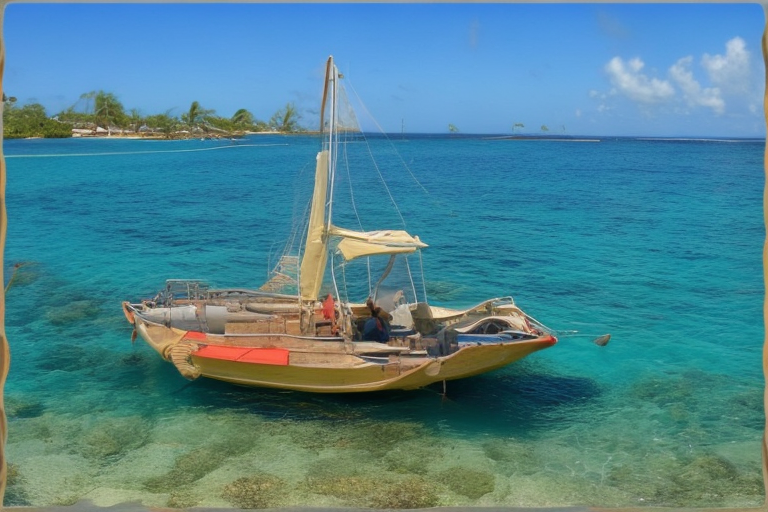}
        \end{subfigure}
        &
        \begin{subfigure}{0.195\textwidth}
            \centering
            \includegraphics[width=\linewidth]{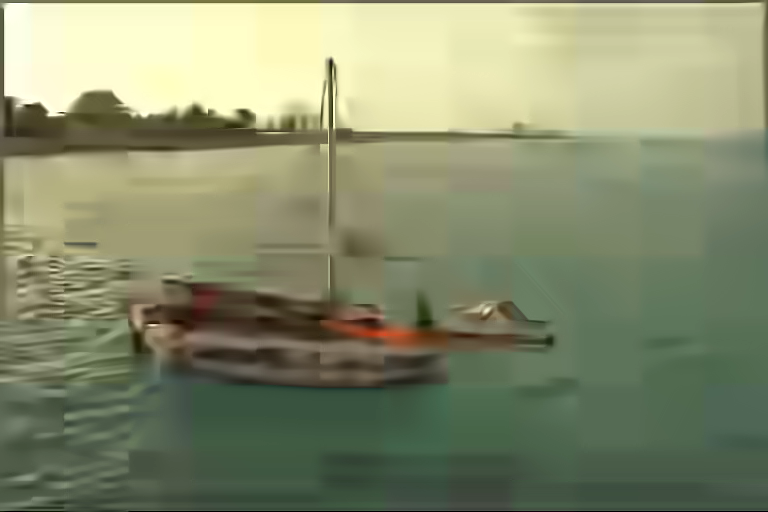}
        \end{subfigure}
        &
        \begin{subfigure}{0.195\textwidth}
            \centering
            \includegraphics[width=\linewidth]{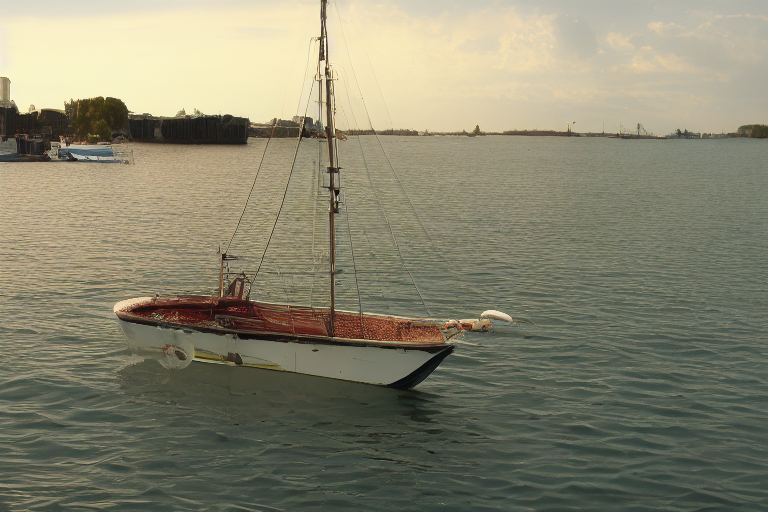}
        \end{subfigure}
        &
        \begin{subfigure}{0.195\textwidth}
            \centering
            \includegraphics[width=\linewidth]{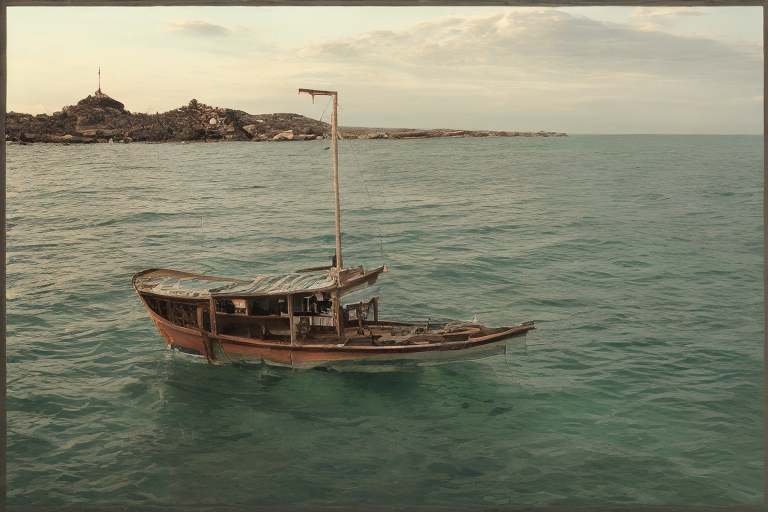}
        \end{subfigure} \\

        \begin{subfigure}{0.195\textwidth}
            \centering
            \includegraphics[width=\linewidth]{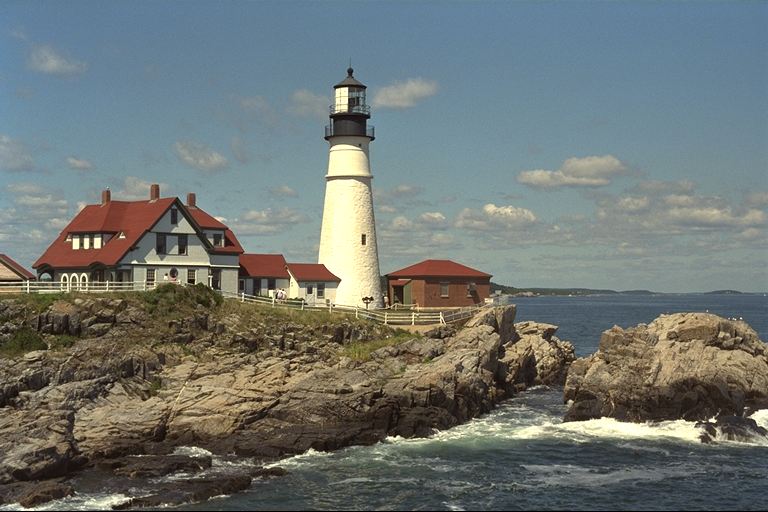}
        \end{subfigure}
        &
        \begin{subfigure}{0.195\textwidth}
            \centering
            \includegraphics[width=\linewidth]{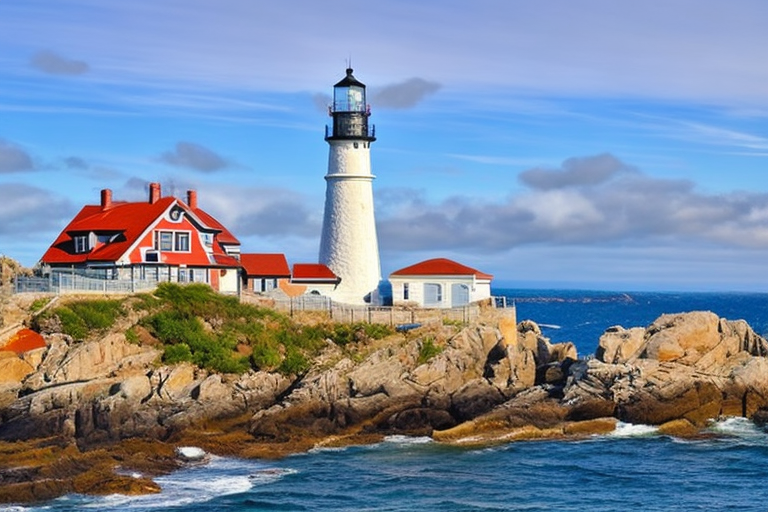}
        \end{subfigure}
        &
        \begin{subfigure}{0.195\textwidth}
            \centering
            \includegraphics[width=\linewidth]{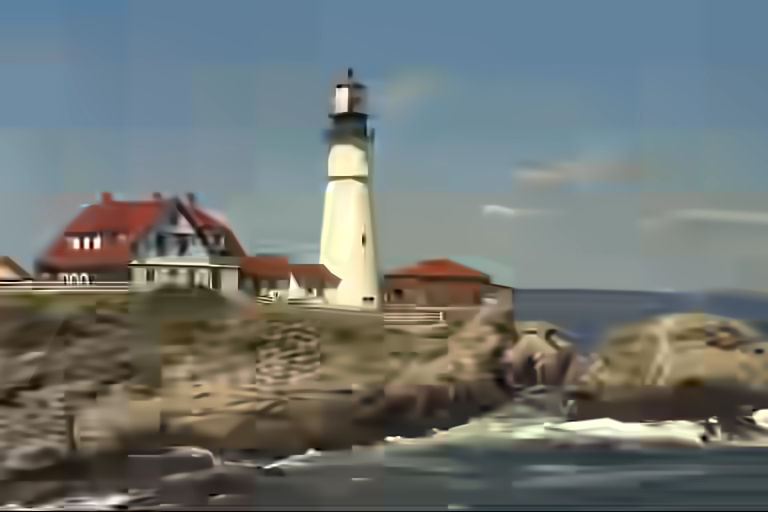}
        \end{subfigure}
        &
        \begin{subfigure}{0.195\textwidth}
            \centering
            \includegraphics[width=\linewidth]{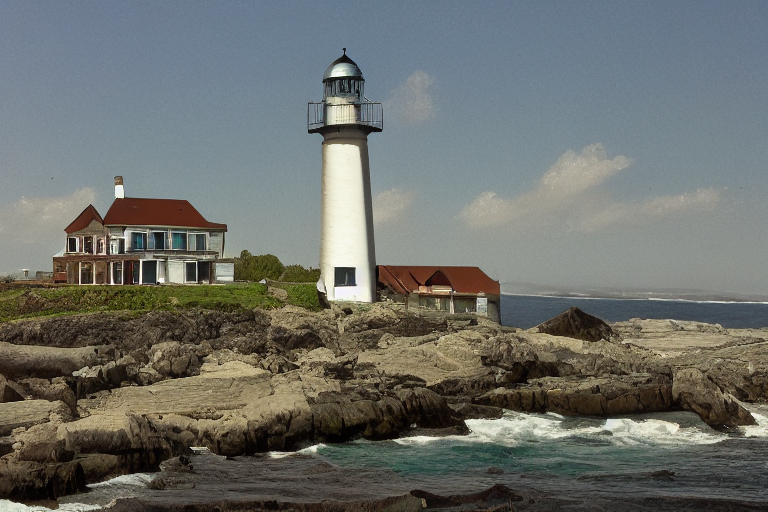}
        \end{subfigure}
        &
        \begin{subfigure}{0.195\textwidth}
            \centering
            \includegraphics[width=\linewidth]{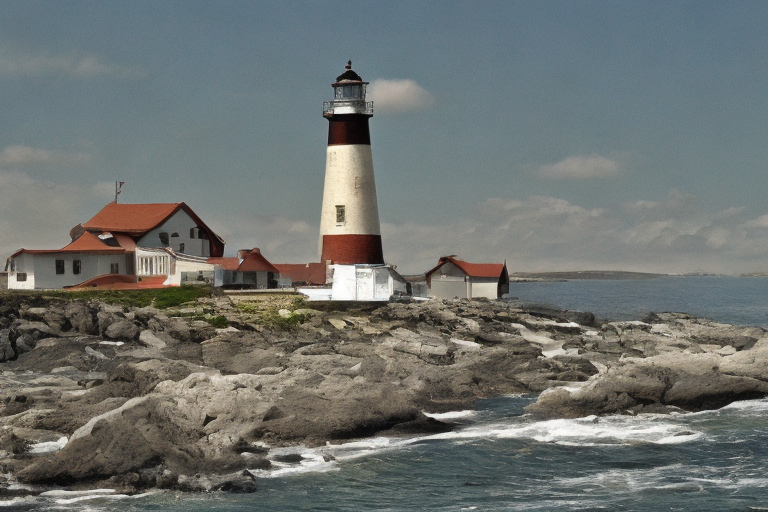}
        \end{subfigure} \\

        \bottomrule

    \end{tabular}

    \caption{Visual comparison of PerCo (SD) to PICS~\cite{lei2023text+sketch}, VTM-20.0, the state-of-the-art non-learned image codec, and PerCo~\cite{careil2024towards}. Notably, PerCo and PerCo (SD) achieve an order of magnitude lower bits per pixel (bpp) compared to competing methods. \textbf{Best viewed electronically.}}
    \label{fig:vis_impressions}
\end{figure*}

To close this gap and to facilitate further research, we introduce PerCo (SD), an open and competitive alternative to PerCo based on the Stable Diffusion architecture~\cite{Rombach_2022_CVPR}, see~\cref{fig:vis_impressions} for visual impressions. In the following, we review the theoretical foundations (\cref{sec:background}), discuss key engineering decisions in adapting PerCo to the Stable Diffusion ecosystem (\cref{sec:challenges}), and provide a comprehensive comparison, both quantitatively and qualitatively (\cref{sec:comparison}).

\section{Background}\label{sec:background}

\textbf{Neural image compression.} Neural image compression uses deep learning/ machine learning techniques to learn compact image representations. This is typically achieved by an auto-encoder-like structure consisting of an encoder $E$ and a decoder $D$, as well as an optional entropy model $P$, which are trained jointly in a data-driven fashion. Specifically, $E$ projects the input image $x$ to a quantized latent representation $y=E(x)$, while $D$ attempts to reverse this process $x'=G(y)$. The learning objective is to minimize the rate-distortion trade-off~\cite{cover2012elements}, with $\lambda > 0$:

\begin{equation}\label{eq:rd_objective}
\mathcal{L}_{RD}=\mathop{\mathbb{E}_{x\sim p_X}}[\lambda r(y) + d(x, x')].
\end{equation}

In~\cref{eq:rd_objective}, the bit-rate is estimated using the cross entropy $r(y)=-\log{P(y)}$, where $P$ represents a probability model of $y$. In practice, an entropy coding method based on $P$ is used to obtain the final bit representation, \eg, using adaptive arithmetic coding. The distortion is measured by a full-reference metric $d(x, x')$ that captures the distance of the reconstruction $x'$ to the original input image $x$. Both terms are weighted by $\lambda$, which enables traversing the rate-distortion curve based on application needs. For a more general overview, we refer the reader to~\cite{CGV-107}.

\textbf{Diffusion models.} Diffusion models~\cite{pmlr-v37-sohl-dickstein15, NEURIPS2020_4c5bcfec} are a type of generative models that approximate the underlying data distribution by learning the inverse of a diffusion process, which is defined as:

\begin{equation}\label{eq:diffusion_process}
q(\mathbf{x}_{1:T}|\mathbf{x}_0) := \prod_{t=1}^T q(\mathbf{x}_t|\mathbf{x}_{t-1}), \quad q(\mathbf{x}_t|\mathbf{x}_{t-1}) := \mathcal{N}(\mathbf{x}_t; \sqrt{1-\beta_t} \mathbf{x}_{t-1}, \beta_t \mathbf{I}).
\end{equation}

In~\cref{eq:diffusion_process}, $q(\mathbf{x}_{1:T}|\mathbf{x}_0)$ denotes the joint distribution of all samples generated across the trajectory of the forward diffusion process in $T$ steps from $\mathbf{x}_1$ up to $\mathbf{x}_T$, given the input image $\mathbf{x}_0$. At each step, Gaussian noise is gradually added to the data following a noise schedule $\beta_t$, such that $q(\mathbf{x}_T|\mathbf{x}_{T-1})=\mathcal{N}(\mathbf{x}_T; \mathbf{0}, \mathbf{I})$. In practice, the forward process can be simulated by $q(\mathbf{x}_t|\mathbf{x}_0) := \mathcal{N}(\mathbf{x}_t; \sqrt{\bar{\alpha}_t} \mathbf{x}_0, (1-\bar{\alpha}_t)\mathbf{I}))$, with $\bar{\alpha}_t=\prod_{s=1}^t (1-\beta_s)$, which enables the convenient parameterization $\mathbf{x}_t = \sqrt{\bar{\alpha}_t} \mathbf{x}_0 + \sqrt{1-\alpha}_t \epsilon$, with $\epsilon \sim \mathcal{N}(\mathbf{0}, \mathbf{I})$.

The reverse diffusion process is defined as:

\begin{equation}\label{eq:diffusion_process_reverse}
p_\theta(\mathbf{x}_{0:T}) := p(\mathbf{x}_T) \prod_{t=1}^T p_\theta(\mathbf{x}_{t-1}|\mathbf{x}_t), \quad p_\theta(\mathbf{x}_{t-1}|\mathbf{x}_t) := \mathcal{N}(\mathbf{x}_{t-1}; \boldsymbol{\mu}_\theta(\mathbf{x}_t, t), \sigma_t^2 \mathbf{I}).
\end{equation}

In~\cref{eq:diffusion_process_reverse}, $p_\theta(\mathbf{x}_{0:T})$ denotes the joint distribution of all samples generated across the trajectory of the reverse diffusion process in $T$ steps from $\mathbf{x}_T$ up to $\mathbf{x}_0$, with $p(\mathbf{x}_T)=q(\mathbf{x}_T|\mathbf{x}_{T-1})=\mathcal{N}(\mathbf{x}_T; \mathbf{0}, \mathbf{I})$, where $p_\theta(\mathbf{x}_{t-1}|\mathbf{x}_t)$ approximates the true denoising distribution $q(\mathbf{x}_{t-1}|\mathbf{x}_t)$ using a parametric Gaussian model (\eg, time-conditional U-Net).

The learning objective is based on the variational lower bound, adapted to the diffusion setting:

\begin{equation}\label{eq:diffusion_learning_objective}
\mathbb{E}_{q(\mathbf{x}_0)}\left[-\log p_\theta(\mathbf{x}_0) \right] \leq \mathbb{E}_{q(\mathbf{x}_0)q(\mathbf{x}_{1:T}|\mathbf{x}_0)} \left[ -\log \frac{p_\theta(\mathbf{x}_{0:T})}{q(\mathbf{x}_{1:T}|\mathbf{x}_0)} \right] := L.
\end{equation}

Ho~\etal~\cite{NEURIPS2020_4c5bcfec} showed that this objective can be further simplified to a noise prediction task, neglecting multiplicative constants, which is widely used in practice and constitutes the foundation of the earlier variants of Stable Diffusion v1.1-v1.5~\cite{Rombach_2022_CVPR}:

\begin{equation}\label{eq:diffusion_learning_objective_simple}
L_\text{simple}(\theta) = \mathbb{E}_{t, \mathbf{x}_0, \epsilon} \left[ \| \epsilon - \epsilon_\theta(\mathbf{x}_t, t) \|^2 \right].
\end{equation}

An extension of~\cref{eq:diffusion_learning_objective_simple} to the conditional case can be achieved by adding side information $z$ (\eg text descriptions of $\mathbf{x}_0$) to the input of the noise prediction network $\epsilon_\theta(\mathbf{x}_t, z, t)$.

\textbf{Latent diffusion models.} Latent diffusion models (LDMs)~\cite{Rombach_2022_CVPR} are a subset of diffusion models that formulate the learning objective~\cref{eq:diffusion_learning_objective} in a latent space (\eg, of a pre-trained auto-encoder), rather than in the pixel space. This change is primarily to reduce the high computational complexity during both training and sampling.

\section{Perceptual compression}\label{sec:challenges}

Perceptual compression, sometimes referred to as generative compression~\cite{Agustsson_2019_ICCV, mentzer2020high} or distribution-preserving compression~\cite{NEURIPS2018_801fd8c2}, extends the traditional rate-distortion objective~\cref{eq:rd_objective} by an additional constraint that forces the reconstructions to follow the underlying data distribution, leading to the rate-distortion-perception trade-off~\cite{ICML-2019-BlauM}.

The key idea of PerCo is to formulate the distortion term $d(x, x')$ in~\cref{eq:rd_objective} within a pre-trained text-conditional LDM, which serves as a powerful generative prior. This type of formulation has recently been also referred to as generative latent coding (as opposed to the regular transform coding paradigm in the pixel space) and is motivated by the fact that the latent space typically has greater sparsity, richer semantics, and better alignment with human perception~\cite{Jia_2024_CVPR}.

\subsection{Model overview}

\begin{figure}[tb]
  \centering
  \includegraphics[height=3.70cm]{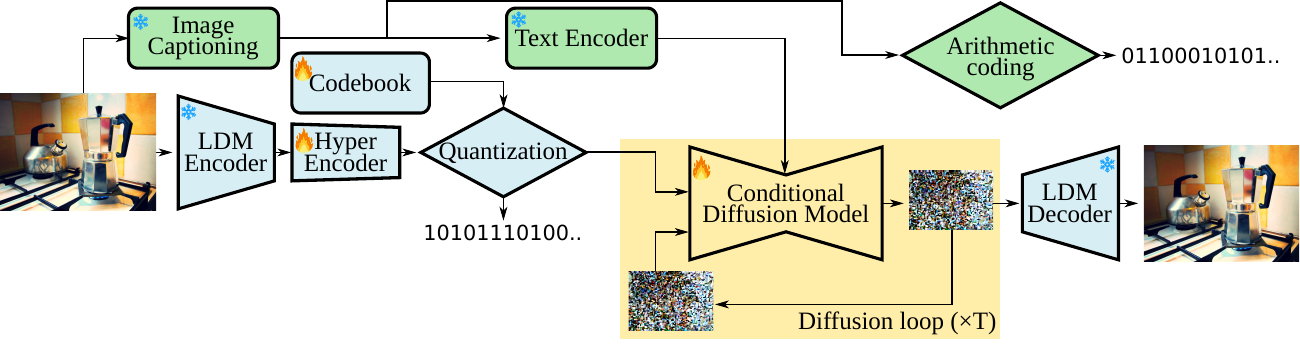}
  \caption{PerCo model overview, adopted from~\cite{careil2024towards}. During training, the hyper-encoder, codebook, and diffusion model are trained, whereas all other components are fixed.
  }
  \label{fig:perco_overview}
\end{figure}

In this section, we provide a short model overview of PerCo (\cref{fig:perco_overview}). The core component is a conditional diffusion model (highlighted yellow) based on a proprietary variant of GLIDE~\cite{pmlr-v162-nichol22a}, which we intend to replace with an open alternative (\cref{subsec:perco_sd}). 

\textbf{Encoding.} To better adapt the LDM to the compression setting, PerCo extracts side information at the encoder side of the form $z=(z_l, z_g)$, where $z_l$ and $z_g$ correspond to local and global features, respectively. In PerCo, $z_l$ corresponds to vector-quantized (VQ) hyper-latent features, extracted by the hyper-encoder, and $z_g$ corresponds to image captions extracted by a pre-trained large language model (BLIP-2~\cite{pmlr-v202-li23q}). Both $z_l$ and $z_g$ are losslessly compressed using arithmetic coding and Lempel-Ziv coding. In PerCo, a uniform coding scheme is used to model $z_l$, \ie the rate term $r(y)$ in~\cref{eq:rd_objective} can be ignored. Various bit-rates can be achieved by using different configurations for the spatial size, denoted by ($h \times w$), and the codebook size $V$: $r(z_l) = \frac{h w \log_2 V}{HW}$ bpp, where ($H \times W$) denotes the input size. The final bit-rate is obtained by $r(z) = r(z_l) + r(z_g)$, where $r(z_g)$ is controlled by the number of tokens ($32$ in the official configuration).

\textbf{Decoding.} At the decoder side, the compressed representations $(z_l, z_g)$ are decoded and subsequently fed into the conditional diffusion model: $z_l$ is upsampled using linear interpolation if required, and spatially concatenated\ with $\mathbf{x}_t$, the input of the first convolution of the denoising network. This is achieved by extending the pre-trained kernel with randomly initialized weights. $z_g$ is passed to a pre-trained text encoder that computes textual embeddings, which are incorporated into the denoising network using cross-attention layers~\cite{NIPS2017_3f5ee243}. 

\subsection{PerCo (SD)}\label{subsec:perco_sd}

Our general goal is to provide an open alternative to PerCo, with ideally highly competitive performance, while following the official design decisions as closely as possible.

\textbf{Challenges.} Among the many available Stable Diffusion options, we choose version 2.1\footnote{\url{https://github.com/Stability-AI/stablediffusion}}, which is similar to the proprietary GLIDE-based LDM, a native v-prediction model~\cite{salimans2022progressive}. Prior to the adoption, we had the following concerns: i) the LDM of Stable Diffusion v2.1 is much smaller than the one used in PerCo. In SD v2.1, we have $866$M, $84$M, and $340$M parameters for the denoising network, auto-encoder, and text-encoder, respectively. PerCo uses a $1.4$B-parameter denoising network ($1.62\times $), a $4.7$B-parameter text encoder ($13.82\times $), and an $83$M-parameter auto-encoder. ii) SD v2.1 by default uses a larger input resolution ($768\times 768$) compared to the target resolution ($512\times 512$). iii) Finally, it remains unclear how the proprietary LDM performs in comparison to existing off-the-shelf models, given the current analysis of the consistency-diversity-realism fronts~\cite{Astolfi2024}. 

\textbf{Core design decisions/ deviations.} In this section, we discuss the core changes over the official configuration. A full detailed comparison is provided in~\cref{tab:comparison}.

\begin{itemize}
    \item Training steps. We limit the number of training steps to $150$k iterations due to resource considerations, which roughly corresponds to $50\%$ of the computation budget of PerCo.
    \item Peak learning rate. We generally find it beneficial to use small learning rates ($1e-5$).
    \item U-Net finetuning. We finetune the whole U-Net, which we find to provide slightly better results. We attribute this observation to an initial resolution/ distribution mismatch.
    \item Extended kernel. We initialize the extended kernel with zeros~\cite{Zhang_2023_ICCV}, which encourages the model to gradually incorporate the additional conditional information ($z_l$).
    \item VQ-module. We additionally $\ell_2$-normalize the codes~\cite{yu2022vectorquantized}, which we find to be crucial to ensure stable training. 
    
\end{itemize}

\begin{table}[tb]
  \caption{Comparison of the design decisions: PerCo (official) vs. PerCo (SD). Key deviations are highlighted in gray and discussed in the main text.}
  \label{tab:comparison}
  \centering
  \begin{tabular}{lll}
    \toprule
          & PerCo (official)     & PerCo (SD) \\
    
    \midrule
    \textbf{Training} &  &      \\
    \cmidrule(r){1-1}
    Training dataset & OpenImagesV6~\cite{OpenImages} ($9$M) & OpenImagesV6~\cite{OpenImages} ($9$M)     \\
    Optimizer & AdamW~\cite{loshchilov2018decoupled}  & AdamW~\cite{loshchilov2018decoupled}    \\
    \rowcolor{verylightgray}
    Training steps & $5$ epochs/ $\approx 300$k  & $150$k ($50\%$)    \\
    \rowcolor{verylightgray}
    Peak learning rate & $1e-4$  & $1e-5$     \\
    Weight decay & $0.01$  & $0.01$     \\
    Linear warm-up & $10$k  & $10$k     \\
    Batch size & $160$ (w/o LPIPS), $40$ w/ LPIPS  & $80$ w/ LPIPS     \\
    \rowcolor{verylightgray}
    U-Net finetuning & linear layers ($15\%$)  & all layers    \\
    LPIPS auxilliary loss & bit-rates $>0.05$bpp  & all bit-rates  \\
    Text conditioning & drop in $10\%$  & drop in $10\%$ \\
    Finetuning grid & $50$ steps  & $1000$ steps (unchanged) \\
    \rowcolor{verylightgray}
    Extended kernel & random initialization  & zero initialization \\
    \rowcolor{verylightgray}
    VQ-module & improved VQ~\cite{yu2022vectorquantized}  & improved VQ~\cite{yu2022vectorquantized} + cosine similarity \\
    \midrule
    \textbf{Inference} &  &      \\
    \cmidrule(r){1-1}
    Scheduler & DDIM~\cite{song2021denoising}  & DDIM~\cite{song2021denoising}     \\
    Denoising steps & $5$ for > $0.05$bpp, else $20$ & $20$  \\
    CFG~\cite{ho2021classifierfree} & $3.0$ & $3.0$     \\
    \bottomrule
  \end{tabular}
\end{table}

\textbf{Further considerations.} We explored finite-scalar quantization (FSQ)~\cite{mentzer2024finite} as a simpler alternative to the sensitive codebook learning paradigm. While FSQ does indeed streamline the training process, it falls short of matching the performance of its VQ counterparts. We further investigated the use of LoRa~\cite{hu2022lora} as an alternative to solely fine-tuning the linear layers to somewhat better quantify the issue of catastrophic forgetting~\cite{8107520}. However, this approach did not yield improved results. Lastly, we explored various $l_z$-conditioning formulations of the diffusion model. We experimented with an additional hyper-decoder, as an alternative to the simple upsample with linear interpolation operation, either by directly using the hyper-decoded local features as input to the diffusion model, or to support auxiliary loss formulations to regularize the hyper-encoder (\eg by enforcing good reconstruction quality of the latent features). In both scenarios, we did not observe additional improvements. This can be partly attributed to the observation that downstream learning tasks yield comparable results in both the latent and pixel spaces~\cite{torfason2018towards}.

\section{Experimental results}\label{sec:comparison}

\textbf{Implementational details.} PerCo (SD) is written in PyTorch~\cite{NEURIPS2019_bdbca288} and built around the diffusers library~\cite{von-platen-etal-2022-diffusers}. As such, PerCo (SD), in general, allows for testing various Stable Diffusion versions (v1, v2) out-of-the-box, with minor adjustments. We use a single DGX H100 system to train all models in a distributed, multi-GPU ($8\times$ H100) setup using full precision. To further accelerate training, all captions are pre-computed and loaded into memory during runtime. PerCo (SD) is also accompanied by a simplified Google Colab demo, which enables training on a single A100-GPU.

\textbf{Evaluation setup.} We adopt the same evaluation protocol as in PerCo~\cite{careil2024towards}. We consider the Kodak~\cite{kodak} and the MSCOCO-30k~\cite{caesar2018cvpr} datasets, which contain $24$ and $30$k images at resolution $512\times768$ and $512\times512$, respectively. We report the FID~\cite{NIPS2017_8a1d6947} and KID~\cite{bińkowski2018demystifying} as a measure of perception, the MS-SSIM~\cite{msssim2003} and LPIPS~\cite{Zhang_2018_CVPR} as a measure of distortion, the CLIP-score~\cite{hessel-etal-2021-clipscore} as a measure of global alignment of reconstructed images and ground truth captions (in PerCo: BLIP 2 generated captions) and finally, the mean intersection over union (mIoU) as a measure of semantic preservation~\cite{schoenfeld2021you}. For more details, we refer the reader to~\cite[Section 4.1 and A Experiment details]{careil2024towards}.

\subsection{Main results}

\begin{figure}[tb]
  \centering
  \includegraphics[width=0.95\linewidth]{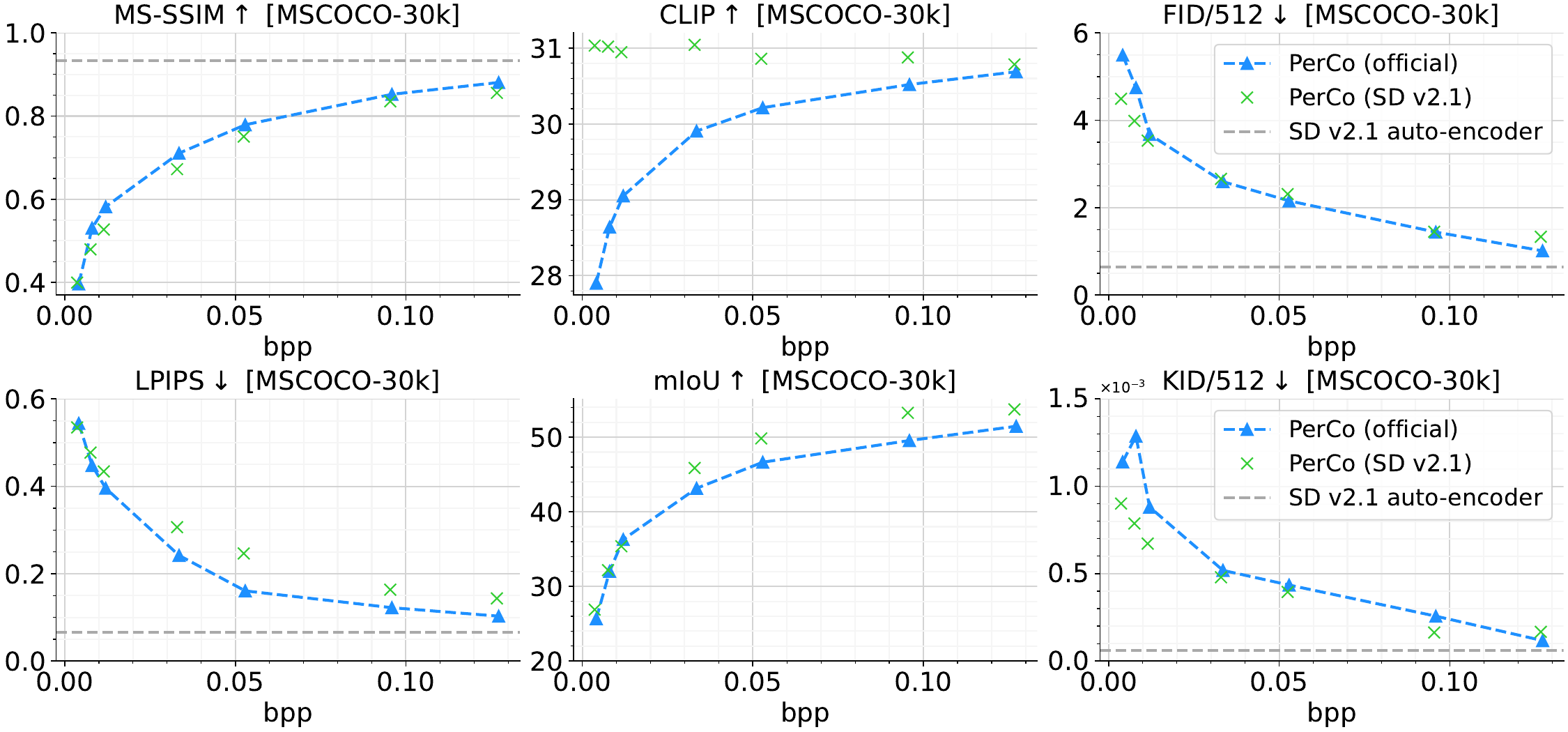}
  \caption{Quantitative comparison: PerCo (official) vs. PerCo (SD)}
  \label{fig:quant_comparison}
\end{figure}

In this section, we quantitatively compare the performance of PerCo (SD v2.1) to the officially reported numbers (\cref{fig:quant_comparison}). All models were trained using a reduced set of optimization steps ($150$k, $50$\% of the official configuration). Note that the performance is bounded by the LDM auto-encoder, denoted as SD v2.1 auto-encoder.

We generally obtain highly competitive results in terms of perception (FID, KID), especially for the ultra-low bit rates. For our lowest bit rate configuration, $0.0036$bpp, we obtain considerably better FID and KID scores compared to PerCo at $0.0041$bpp ($4.49$ vs. $5.49$ and $0.0009$ vs. $0.0011$). This benefit comes, however, at the cost of consistently lower image fidelity (MS-SSIM, LPIPS). Besides the notorious rate-distortion-perception trade-off~\cite{ICML-2019-BlauM}, we attribute this gap to the different model capacities being used (LDM $866$M vs. $1.4$B, Text encoder $340$M vs. $4.7$B).  Intuitively, PerCo attempts to recover the latent image code from only a short text description and vector-quantized hyper-latent features, which arguably requires a sophisticated generative prior. We further obtain superior CLIP and mIoU scores. PerCo (SD) tends, however, to use slightly shorter, perhaps more generic text descriptions ($0.00165$bpp vs. $0.0022$bpp) due to presumably different BLIP 2 configurations. As such, the CLIP scores might not be directly comparable. In our case, the CLIP scores also seem less dependent on the bit rate.

Finally, it is worth mentioning that we did not apply post-hoc filtering methods~\cite{karthik2023if} to further boost performance. Like all probabilistic methods, PerCo (SD) is sensitive to the initial random seed. Therefore, future work should report the mean and standard deviation across multiple test runs.

\subsection{Ablations/ further results}

Both PerCo and PerCo (SD) rely on the DDIM scheduler. We find that the default configuration remains a good choice (classifier-free guidance scale of $3$ and $20$ sampling steps). For additional details and further results, see~\cref{appendix}.

\section{Related work}

We limit this section to concurrent approaches for ultra-low bit-rate image compression that leverage powerful pre-trained foundation models and refer the reader for a broader overview to~\cite[Related work]{careil2024towards}. Conditioning modalities explored in these methods include prompt inversion and compressed sketches~\cite{wen2023hard, lei2023text+sketch}, text descriptions obtained by a commercial large language model (GPT-4 Vision~\cite{GPT4Vision}), semantic label maps and compressed image features~\cite{li2024misc}, CLIP image features and color palettes~\cite{pmlr-v139-radford21a, bachard:hal-04478601}, and textual inversion combined with a variant of classifier guidance, dubbed compression guidance~\cite{gal2023an, dhariwal2021diffusion, pan_2022}. Relic~\etal~\cite{Relic2024} takes a slightly different approach by treating the removal of quantization error as the denoising task, aiming to recover lost information in the transmitted image latent. In all cases, some form of Stable Diffusion~\cite{Rombach_2022_CVPR} is used (ControlNet~\cite{Zhang_2023_ICCV}, DiffBIR~\cite{lin2024diffbir} and Stable unCLIP~\cite{Rombach_2022_CVPR}), with no changes to the official weights.

\section{Conclusion}\label{sec:conclusion}
In this paper, we introduced PerCo (SD), an open and competitive alternative to PerCo, the current state-of-the-art for ultra-low bit-rate image compression. We revisited the theoretical foundations, described our engineering efforts in translating PerCo to the Stable Diffusion ecosystem, and provided an in-depth analysis of both approaches. We hope our work contributes to a deeper understanding of the underlying mechanisms and paves the way for future advancements in the field. Code and models will be released at \url{https://github.com/Nikolai10/PerCo}.

\textbf{Limitations.} PerCo (SD) inherits the limitations described in the original work. In its current state, PerCo (SD) can only handle medium-sized images (\eg, $512 \times 512$). Possible solutions have been discussed in~\cite[section 5]{torfason2018towards}. Finally, PerCo (SD) is based on a much smaller LDM ($866$M vs. $1.4$B) - we leave the exploration of more powerful foundation models for future work.

\begin{ack}
This work was supported by the German Federal Ministry of Education and Research under the funding program Forschung an Fachhochschulen - FKZ 13FH019KI2. The authors would further like to thank Marlène Careil for helpful insights and evaluation data.
\end{ack}

\bibliographystyle{plain}
\bibliography{egbib}

\begin{thebibliography}{10}

\bibitem{Agustsson_2019_ICCV}
Eirikur Agustsson, Michael Tschannen, Fabian Mentzer, Radu Timofte, and Luc~Van Gool.
\newblock Generative adversarial networks for extreme learned image compression.
\newblock In {\em Proceedings of the IEEE/CVF International Conference on Computer Vision (ICCV)}, 2019.

\bibitem{Astolfi2024}
P.~Astolfi, M.~Careil, M.~Hall, O.~Mañas, M.~Muckley, J.~Verbeek, A.~R. Soriano, and M.~Drozdzal.
\newblock Consistency-diversity-realism pareto fronts of conditional image generative models.
\newblock {\em arXiv: 2406.10429}, 2024.

\bibitem{bachard:hal-04478601}
Tom Bachard, Tom Bordin, and Thomas Maugey.
\newblock {CoCliCo: Extremely low bitrate image compression based on CLIP semantic and tiny color map}.
\newblock In {\em {PCS 2024 - Picture Coding Symposium}}, 2024.

\bibitem{bińkowski2018demystifying}
Mikołaj Bińkowski, Dougal~J. Sutherland, Michael Arbel, and Arthur Gretton.
\newblock Demystifying {MMD} {GAN}s.
\newblock In {\em International Conference on Learning Representations}, 2018.

\bibitem{ICML-2019-BlauM}
Yochai Blau and Tomer Michaeli.
\newblock {Rethinking Lossy Compression: The Rate-Distortion-Perception Tradeoff}.
\newblock In {\em {Proceedings of the 36th International Conference on Machine Learning}}, 2019.

\bibitem{Bommasani2021FoundationModels_short}
Rishi Bommasani et~al.
\newblock On the opportunities and risks of foundation models.
\newblock {\em arXiv: 2108.07258}, 2021.

\bibitem{caesar2018cvpr}
Holger Caesar, Jasper Uijlings, and Vittorio Ferrari.
\newblock Coco-stuff: Thing and stuff classes in context.
\newblock In {\em Proceedings of the IEEE Conference on Computer Vision and Pattern Recognition (CVPR)}, 2018.

\bibitem{careil2024towards}
Marlene Careil, Matthew~J. Muckley, Jakob Verbeek, and St{\'e}phane Lathuili{\`e}re.
\newblock Towards image compression with perfect realism at ultra-low bitrates.
\newblock In {\em The Twelfth International Conference on Learning Representations}, 2024.

\bibitem{cover2012elements}
Thomas~M Cover and Joy~A Thomas.
\newblock {\em {Elements of information theory}}.
\newblock John Wiley \& Sons, 2012.

\bibitem{dhariwal2021diffusion}
Prafulla Dhariwal and Alexander~Quinn Nichol.
\newblock Diffusion models beat {GAN}s on image synthesis.
\newblock In {\em Advances in Neural Information Processing Systems}, 2021.

\bibitem{gal2023an}
Rinon Gal, Yuval Alaluf, Yuval Atzmon, Or~Patashnik, Amit~Haim Bermano, Gal Chechik, and Daniel Cohen-or.
\newblock An image is worth one word: Personalizing text-to-image generation using textual inversion.
\newblock In {\em The Eleventh International Conference on Learning Representations}, 2023.

\bibitem{NIPS2014_5ca3e9b1}
Ian Goodfellow, Jean Pouget-Abadie, Mehdi Mirza, Bing Xu, David Warde-Farley, Sherjil Ozair, Aaron Courville, and Yoshua Bengio.
\newblock Generative adversarial nets.
\newblock In {\em Advances in Neural Information Processing Systems}, 2014.

\bibitem{hessel-etal-2021-clipscore}
Jack Hessel, Ari Holtzman, Maxwell Forbes, Ronan Le~Bras, and Yejin Choi.
\newblock {CLIPS}core: A reference-free evaluation metric for image captioning.
\newblock In {\em Proceedings of the 2021 Conference on Empirical Methods in Natural Language Processing}, 2021.

\bibitem{NIPS2017_8a1d6947}
Martin Heusel, Hubert Ramsauer, Thomas Unterthiner, Bernhard Nessler, and Sepp Hochreiter.
\newblock Gans trained by a two time-scale update rule converge to a local nash equilibrium.
\newblock In {\em Advances in Neural Information Processing Systems}, 2017.

\bibitem{NEURIPS2020_4c5bcfec}
Jonathan Ho, Ajay Jain, and Pieter Abbeel.
\newblock Denoising diffusion probabilistic models.
\newblock In {\em Advances in Neural Information Processing Systems}, 2020.

\bibitem{ho2021classifierfree}
Jonathan Ho and Tim Salimans.
\newblock Classifier-free diffusion guidance.
\newblock In {\em NeurIPS 2021 Workshop on Deep Generative Models and Downstream Applications}, 2021.

\bibitem{hu2022lora}
Edward~J Hu, yelong shen, Phillip Wallis, Zeyuan Allen-Zhu, Yuanzhi Li, Shean Wang, Lu~Wang, and Weizhu Chen.
\newblock Lo{RA}: Low-rank adaptation of large language models.
\newblock In {\em International Conference on Learning Representations}, 2022.

\bibitem{Jia_2024_CVPR}
Zhaoyang Jia, Jiahao Li, Bin Li, Houqiang Li, and Yan Lu.
\newblock Generative latent coding for ultra-low bitrate image compression.
\newblock In {\em Proceedings of the IEEE/CVF Conference on Computer Vision and Pattern Recognition (CVPR)}, 2024.

\bibitem{karthik2023if}
Shyamgopal Karthik, Karsten Roth, Massimiliano Mancini, and Zeynep Akata.
\newblock If at first you don't succeed, try, try again: Faithful diffusion-based text-to-image generation by selection.
\newblock {\em arXiv: 2305.13308}, 2023.

\bibitem{kodak}
Eastman Kodak.
\newblock Kodak lossless true color image suite ({PhotoCD PCD0992}).

\bibitem{OpenImages}
Alina Kuznetsova, Hassan Rom, Neil Alldrin, Jasper Uijlings, Ivan Krasin, Jordi Pont-Tuset, Shahab Kamali, Stefan Popov, Matteo Malloci, Alexander Kolesnikov, Tom Duerig, and Vittorio Ferrari.
\newblock The open images dataset v4: Unified image classification, object detection, and visual relationship detection at scale.
\newblock {\em IJCV}, 2020.

\bibitem{lei2023text+sketch}
Eric Lei, Yi\u{g}it~Berkay Uslu, Hamed Hassani, and Shirin~Saeedi Bidokhti.
\newblock Text+ sketch: Image compression at ultra low rates.
\newblock In {\em ICML 2023 Workshop on Neural Compression: From Information Theory to Applications}, 2023.

\bibitem{li2024misc}
Chunyi Li, Guo Lu, Donghui Feng, Haoning Wu, Zicheng Zhang, Xiaohong Liu, Guangtao Zhai, Weisi Lin, and Wenjun Zhang.
\newblock Misc: Ultra-low bitrate image semantic compression driven by large multimodal model.
\newblock {\em arXiv: 2402.16749}, 2024.

\bibitem{pmlr-v202-li23q}
Junnan Li, Dongxu Li, Silvio Savarese, and Steven Hoi.
\newblock {BLIP}-2: Bootstrapping language-image pre-training with frozen image encoders and large language models.
\newblock In {\em Proceedings of the 40th International Conference on Machine Learning}, 2023.

\bibitem{8107520}
Zhizhong Li and Derek Hoiem.
\newblock Learning without forgetting.
\newblock {\em IEEE Transactions on Pattern Analysis and Machine Intelligence}, 2018.

\bibitem{lin2024diffbir}
Xinqi Lin, Jingwen He, Ziyan Chen, Zhaoyang Lyu, Bo~Dai, Fanghua Yu, Wanli Ouyang, Yu~Qiao, and Chao Dong.
\newblock Diffbir: Towards blind image restoration with generative diffusion prior.
\newblock {\em arXiv: 2308.15070}, 2024.

\bibitem{loshchilov2018decoupled}
Ilya Loshchilov and Frank Hutter.
\newblock Decoupled weight decay regularization.
\newblock In {\em International Conference on Learning Representations}, 2019.

\bibitem{mentzer2024finite}
Fabian Mentzer, David Minnen, Eirikur Agustsson, and Michael Tschannen.
\newblock Finite scalar quantization: {VQ}-{VAE} made simple.
\newblock In {\em The Twelfth International Conference on Learning Representations}, 2024.

\bibitem{mentzer2020high}
Fabian Mentzer, George~D Toderici, Michael Tschannen, and Eirikur Agustsson.
\newblock High-fidelity generative image compression.
\newblock {\em Advances in Neural Information Processing Systems}, 2020.

\bibitem{pmlr-v162-nichol22a}
Alexander~Quinn Nichol, Prafulla Dhariwal, Aditya Ramesh, Pranav Shyam, Pamela Mishkin, Bob Mcgrew, Ilya Sutskever, and Mark Chen.
\newblock {GLIDE}: Towards photorealistic image generation and editing with text-guided diffusion models.
\newblock In {\em Proceedings of the 39th International Conference on Machine Learning}, 2022.

\bibitem{GPT4Vision}
OpenAI, Josh Achiam, et~al.
\newblock Gpt-4 technical report.
\newblock {\em arXiv: 2303.08774}, 2024.

\bibitem{pan_2022}
Zhihong Pan, Xin Zhou, and Hao Tian.
\newblock {Extreme generative image compression by learning text embedding from diffusion models}.
\newblock {\em arXiv: 2211.07793}, 2022.

\bibitem{NEURIPS2019_bdbca288}
Adam Paszke et~al.
\newblock Pytorch: An imperative style, high-performance deep learning library.
\newblock In {\em Advances in Neural Information Processing Systems}, 2019.

\bibitem{pmlr-v139-radford21a}
Alec Radford, Jong~Wook Kim, Chris Hallacy, Aditya Ramesh, Gabriel Goh, Sandhini Agarwal, Girish Sastry, Amanda Askell, Pamela Mishkin, Jack Clark, Gretchen Krueger, and Ilya Sutskever.
\newblock Learning transferable visual models from natural language supervision.
\newblock In {\em Proceedings of the 38th International Conference on Machine Learning}, 2021.

\bibitem{Relic2024}
Lucas Relic, Roberto Azevedo, Markus Gross, and Christopher Schroers.
\newblock Lossy image compression with foundation diffusion models.
\newblock {\em arXiv: 2303.08774}, 2024.

\bibitem{Rombach_2022_CVPR}
Robin Rombach, Andreas Blattmann, Dominik Lorenz, Patrick Esser, and Bj\"orn Ommer.
\newblock High-resolution image synthesis with latent diffusion models.
\newblock In {\em Proceedings of the IEEE/CVF Conference on Computer Vision and Pattern Recognition (CVPR)}, 2022.

\bibitem{salimans2022progressive}
Tim Salimans and Jonathan Ho.
\newblock Progressive distillation for fast sampling of diffusion models.
\newblock In {\em International Conference on Learning Representations}, 2022.

\bibitem{schoenfeld2021you}
Edgar Sch{\"o}nfeld, Vadim Sushko, Dan Zhang, Juergen Gall, Bernt Schiele, and Anna Khoreva.
\newblock You only need adversarial supervision for semantic image synthesis.
\newblock In {\em International Conference on Learning Representations}, 2021.

\bibitem{pmlr-v37-sohl-dickstein15}
Jascha Sohl-Dickstein, Eric Weiss, Niru Maheswaranathan, and Surya Ganguli.
\newblock Deep unsupervised learning using nonequilibrium thermodynamics.
\newblock In {\em Proceedings of the 32nd International Conference on Machine Learning}, 2015.

\bibitem{song2021denoising}
Jiaming Song, Chenlin Meng, and Stefano Ermon.
\newblock Denoising diffusion implicit models.
\newblock In {\em International Conference on Learning Representations}, 2021.

\bibitem{torfason2018towards}
Róbert Torfason, Fabian Mentzer, Eiríkur Ágústsson, Michael Tschannen, Radu Timofte, and Luc~Van Gool.
\newblock Towards image understanding from deep compression without decoding.
\newblock In {\em International Conference on Learning Representations}, 2018.

\bibitem{NEURIPS2018_801fd8c2}
Michael Tschannen, Eirikur Agustsson, and Mario Lucic.
\newblock Deep generative models for distribution-preserving lossy compression.
\newblock In {\em Advances in Neural Information Processing Systems}, 2018.

\bibitem{NIPS2017_3f5ee243}
Ashish Vaswani, Noam Shazeer, Niki Parmar, Jakob Uszkoreit, Llion Jones, Aidan~N Gomez, \L~ukasz Kaiser, and Illia Polosukhin.
\newblock Attention is all you need.
\newblock In {\em Advances in Neural Information Processing Systems}, 2017.

\bibitem{von-platen-etal-2022-diffusers}
Patrick von Platen, Suraj Patil, Anton Lozhkov, Pedro Cuenca, Nathan Lambert, Kashif Rasul, Mishig Davaadorj, Dhruv Nair, Sayak Paul, William Berman, Yiyi Xu, Steven Liu, and Thomas Wolf.
\newblock Diffusers: State-of-the-art diffusion models.
\newblock \url{https://github.com/huggingface/diffusers}, 2022.

\bibitem{msssim2003}
Zhou Wang, Eero~P Simoncelli, and Alan~C Bovik.
\newblock Multiscale structural similarity for image quality assessment.
\newblock In {\em The Thrity-Seventh Asilomar Conference on Signals, Systems \& Computers, 2003}, 2003.

\bibitem{wen2023hard}
Yuxin Wen, Neel Jain, John Kirchenbauer, Micah Goldblum, Jonas Geiping, and Tom Goldstein.
\newblock Hard prompts made easy: Gradient-based discrete optimization for prompt tuning and discovery.
\newblock In {\em Thirty-seventh Conference on Neural Information Processing Systems}, 2023.

\bibitem{CGV-107}
Yibo Yang, Stephan Mandt, and Lucas Theis.
\newblock An introduction to neural data compression.
\newblock {\em Foundations and Trends® in Computer Graphics and Vision}, 2023.

\bibitem{yu2022vectorquantized}
Jiahui Yu, Xin Li, Jing~Yu Koh, Han Zhang, Ruoming Pang, James Qin, Alexander Ku, Yuanzhong Xu, Jason Baldridge, and Yonghui Wu.
\newblock Vector-quantized image modeling with improved {VQGAN}.
\newblock In {\em International Conference on Learning Representations}, 2022.

\bibitem{Zhang_2023_ICCV}
Lvmin Zhang, Anyi Rao, and Maneesh Agrawala.
\newblock Adding conditional control to text-to-image diffusion models.
\newblock In {\em Proceedings of the IEEE/CVF International Conference on Computer Vision (ICCV)}, 2023.

\bibitem{Zhang_2018_CVPR}
Richard Zhang, Phillip Isola, Alexei~A. Efros, Eli Shechtman, and Oliver Wang.
\newblock The unreasonable effectiveness of deep features as a perceptual metric.
\newblock In {\em Proceedings of the IEEE Conference on Computer Vision and Pattern Recognition (CVPR)}, 2018.

\end{thebibliography}


\appendix

\section{Appendix / supplemental material}\label{appendix}

\subsection{Additional quantitative results}

\begin{figure}[h]
  \centering
  \includegraphics[width=0.49\linewidth]{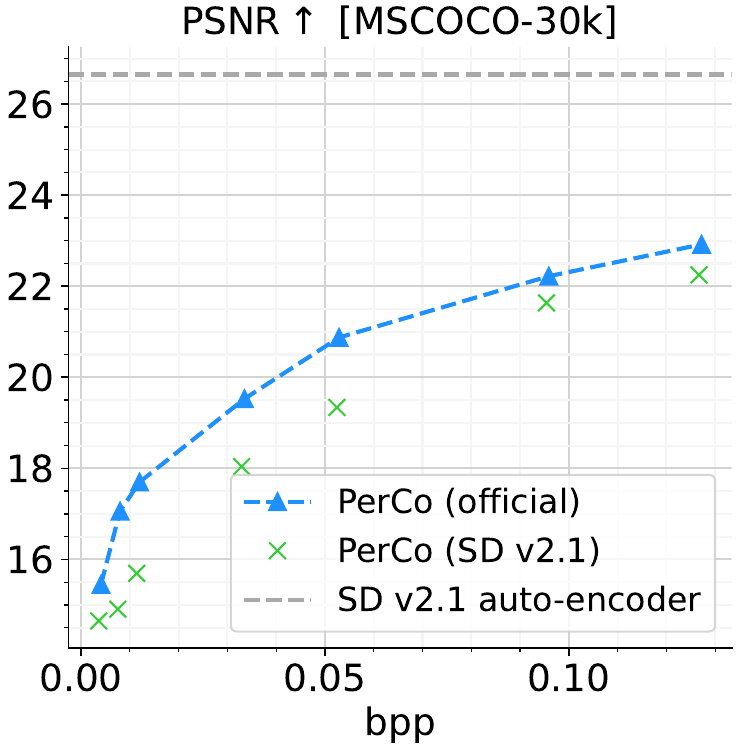}
  \caption{Quantitative comparison on the MSCOCO-30k dataset: PerCo (official) vs. PerCo (SD). We have not tried to tune our model towards better PSNR scores, as these low-level distortion metrics are known to be less meaningful for low rates~\cite{careil2024towards}.}
  \label{fig:perf_coco_psnr}
\end{figure}

\begin{figure}[tb]
  \centering
  \includegraphics[width=0.49\linewidth]{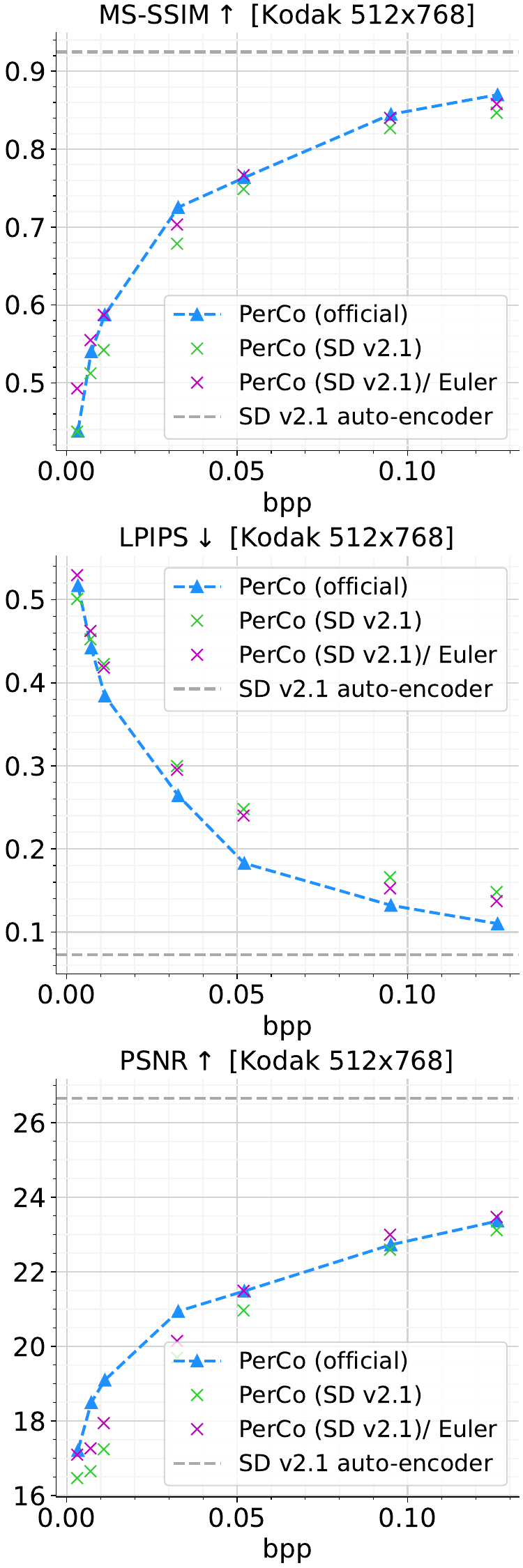}
  \caption{Quantitative comparison on the Kodak dataset: PerCo (official) vs. PerCo (SD). We further show another model configuration based on the \texttt{EulerAncestralDiscreteScheduler}, which we found to produce consistently lower distortion at the cost of, however, slightly decreased perceptual characteristics. Note that the PerCo (SD) performance is bounded by the auto-encoder.}
  \label{fig:perf_kodak}
\end{figure}

\textbf{More results.} We provide additional quantitative results on the MSCOC0-30k and Kodak datasets in~\cref{fig:perf_coco_psnr} and~\cref{fig:perf_kodak}, respectively. We observe characteristics similar to those of the main results. 

\textbf{Inference speed.} We refer the reader to~\cite[A Experimental details, Inference Speed]{careil2024towards}. The encoder speed of PerCo (SD) is supposed to be identical to PerCo, as the only difference lies within the LDM, which runs on the decoder side. As the LDM in PerCo (SD) is considerably smaller ($866$M vs. $1.4$B) and based on a similar architecture, we assume that the decoder speed of PerCo (SD) is at least comparable to PerCo.

\subsection{Additional visual results}

\begin{figure*}[ht]
    \setlength{\tabcolsep}{1.0pt}  
    \renewcommand{\arraystretch}{1.0}  
    \centering
    \scriptsize
    \begin{tabular}{cc}
        \toprule
        Original & PICS ($0.0025$bpp) \\
        \midrule
        \begin{subfigure}{0.49\textwidth}
            \centering
            \includegraphics[width=\linewidth]{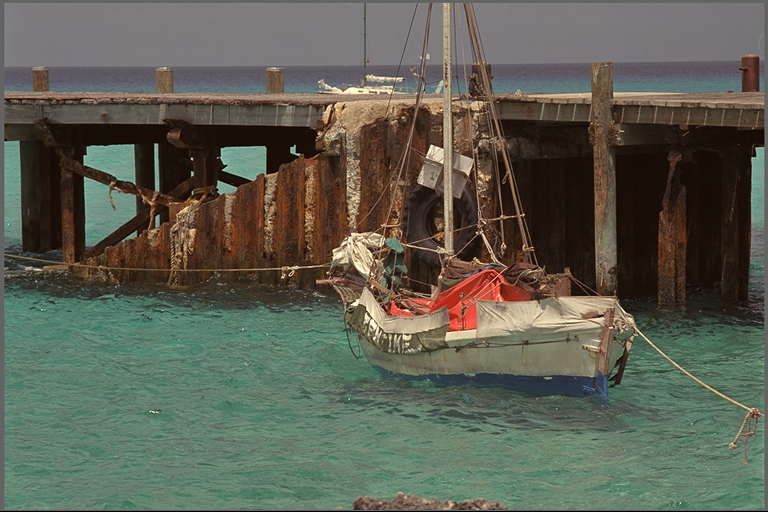}
        \end{subfigure}
        &
        \begin{subfigure}{0.49\textwidth}
            \centering
            \includegraphics[width=\linewidth]{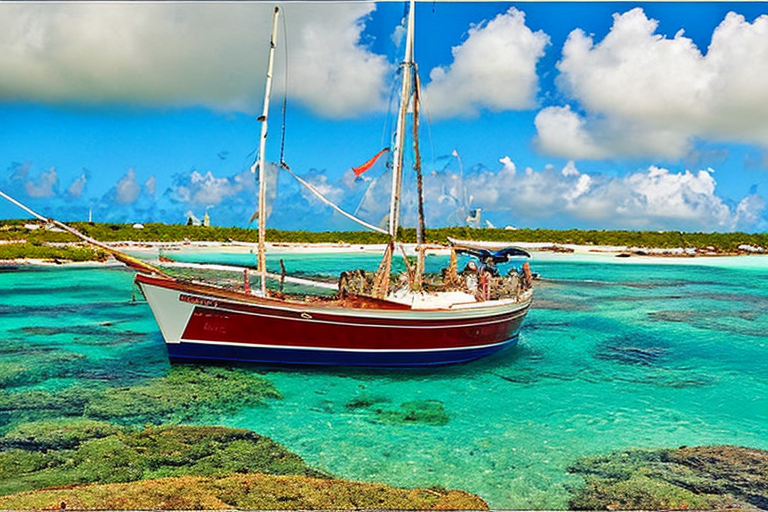}
        \end{subfigure}\\
        
        VTM-20.0 ($0.025$bpp) & MS-ILLM ($0.0065$bpp) \\
        \begin{subfigure}{0.49\textwidth}
            \centering
            \includegraphics[width=\linewidth]{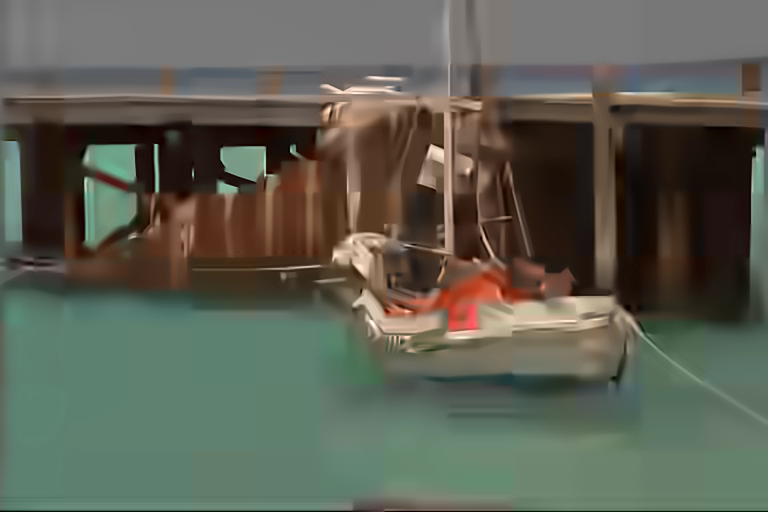}
        \end{subfigure}
        &
        \begin{subfigure}{0.49\textwidth}
            \centering
            \includegraphics[width=\linewidth]{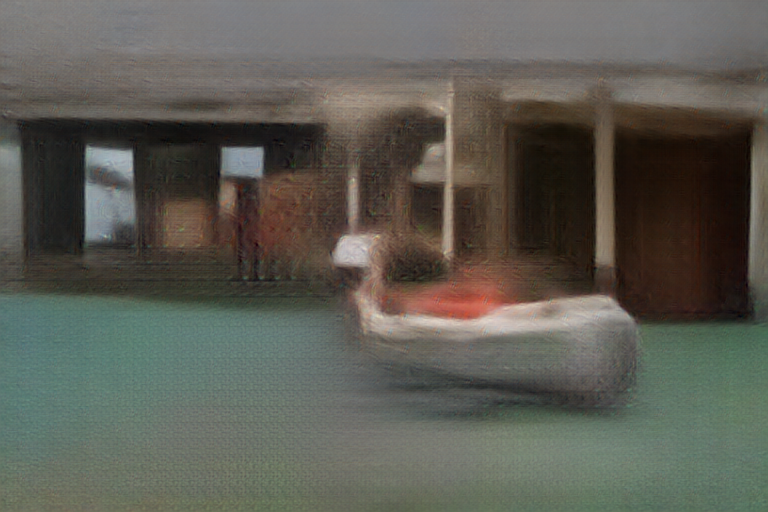}
        \end{subfigure}\\
        
        PerCo ($0.0032$bpp) & Ours ($0.0031$bpp) \\
        \begin{subfigure}{0.49\textwidth}
            \centering
            \includegraphics[width=\linewidth]{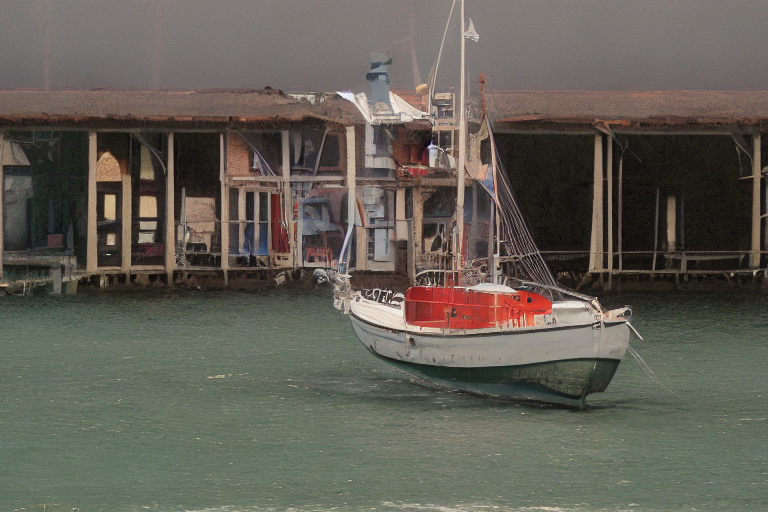}
        \end{subfigure}
        &
        \begin{subfigure}{0.49\textwidth}
            \centering
            \includegraphics[width=\linewidth]{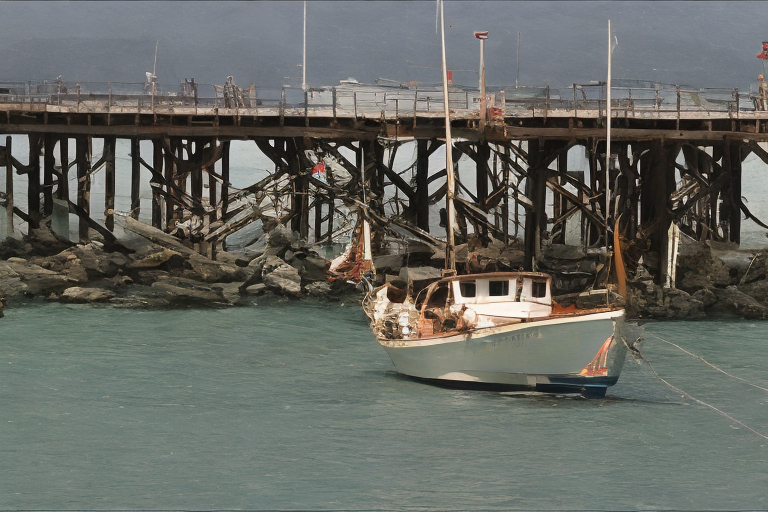}
        \end{subfigure}\\
        
        \bottomrule

    \end{tabular}

    \caption{Visual comparison of PerCo (SD) to PICS~\cite{lei2023text+sketch}, VTM-20.0, the state-of-the-art non-learned image codec, MS-ILLM (Muckley~\etal ICML 2023), and PerCo~\cite{careil2024towards}.}
    \label{fig:vis_impressions_1}
\end{figure*}

\begin{figure*}[ht]
    \setlength{\tabcolsep}{1.0pt}  
    \renewcommand{\arraystretch}{1.0}  
    \centering
    \scriptsize
    \begin{tabular}{cc}
        \toprule
        Original & PICS ($0.028$bpp) \\
        \midrule
        \begin{subfigure}{0.49\textwidth}
            \centering
            \includegraphics[width=\linewidth]{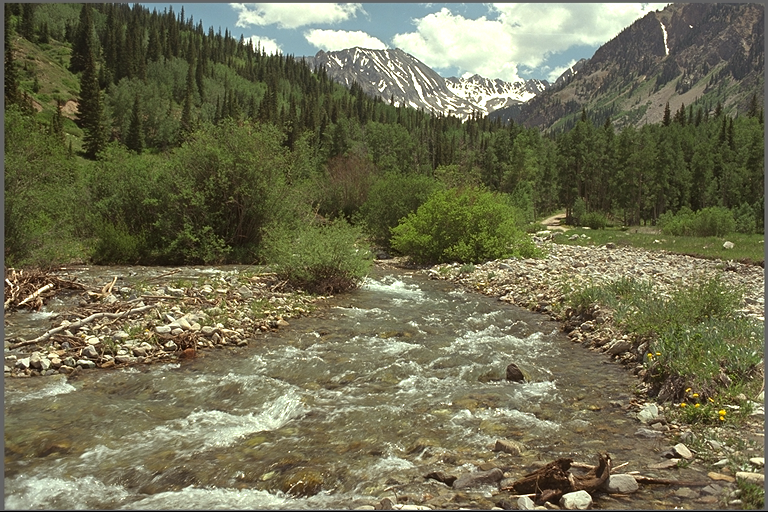}
        \end{subfigure}
        &
        \begin{subfigure}{0.49\textwidth}
            \centering
            \includegraphics[width=\linewidth]{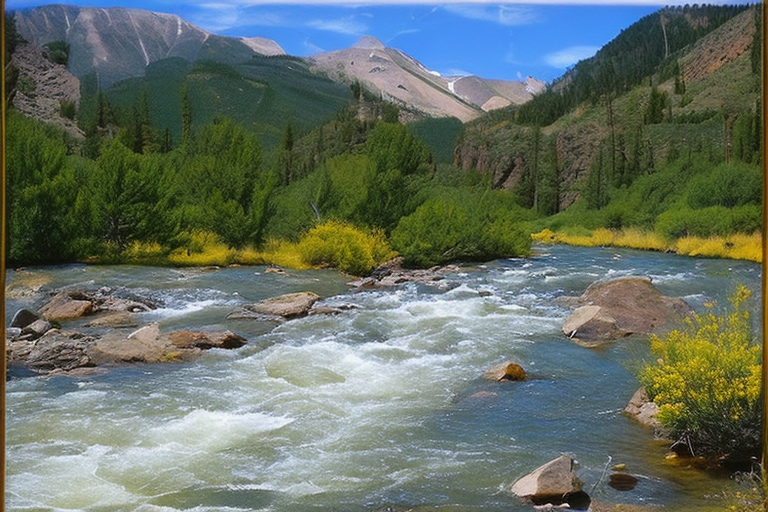}
        \end{subfigure}\\
        
        VTM-20.0 ($0.025$bpp) & MS-ILLM ($ 0.013$bpp) \\
        \begin{subfigure}{0.49\textwidth}
            \centering
            \includegraphics[width=\linewidth]{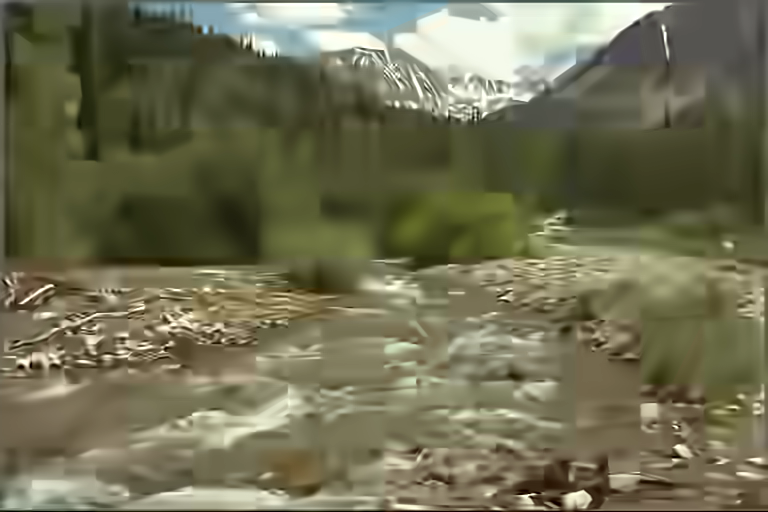}
        \end{subfigure}
        &
        \begin{subfigure}{0.49\textwidth}
            \centering
            \includegraphics[width=\linewidth]{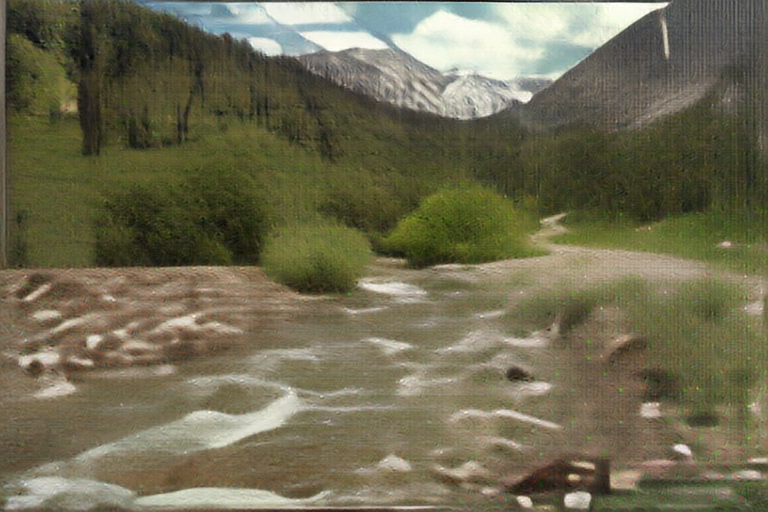}
        \end{subfigure}\\
        
        PerCo ($0.011$bpp) & Ours ($0.011$bpp) \\
        \begin{subfigure}{0.49\textwidth}
            \centering
            \includegraphics[width=\linewidth]{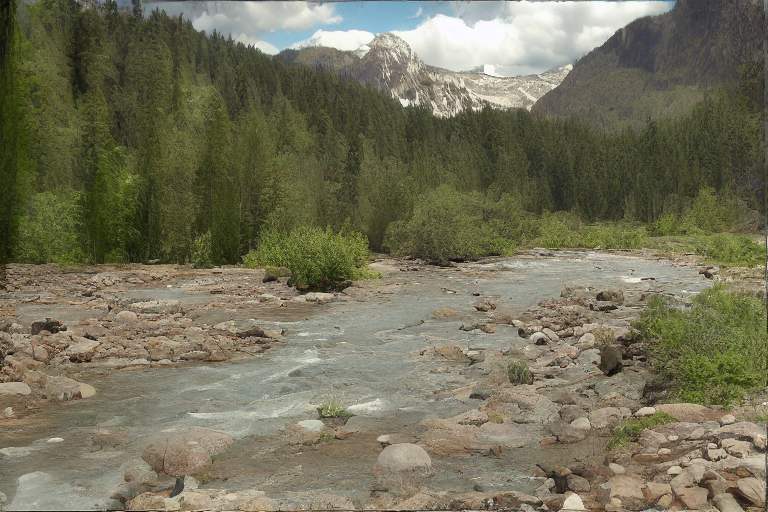}
        \end{subfigure}
        &
        \begin{subfigure}{0.49\textwidth}
            \centering
            \includegraphics[width=\linewidth]{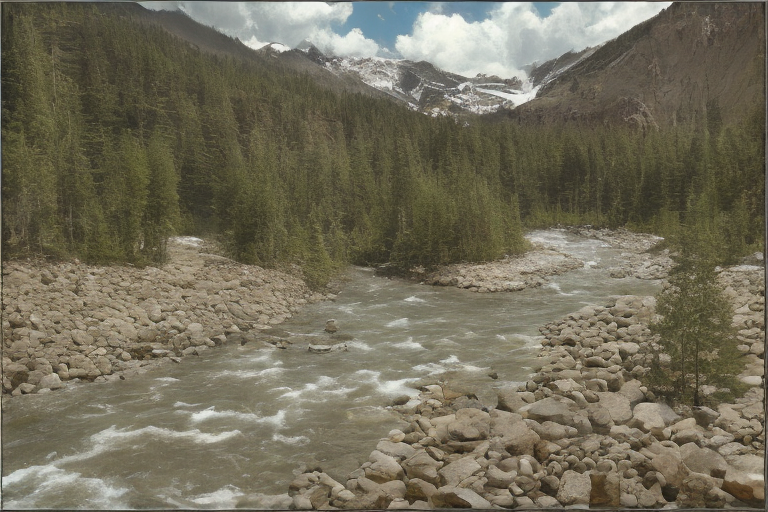}
        \end{subfigure}\\
        
        \bottomrule

    \end{tabular}

    \caption{Visual comparison of PerCo (SD) to PICS~\cite{lei2023text+sketch}, VTM-20.0, the state-of-the-art non-learned image codec, MS-ILLM (Muckley~\etal ICML 2023), and PerCo~\cite{careil2024towards}.}
    \label{fig:vis_impressions_2}
\end{figure*}

\textbf{Additional visual comparisons.} We provide additional visual results in~\cref{fig:vis_impressions_1} and~\cref{fig:vis_impressions_2}. We find that PerCo (SD) produces pleasing reconstructions that are comparable to PerCo.

\begin{figure*}[ht]
    \setlength{\tabcolsep}{1.0pt}  
    \renewcommand{\arraystretch}{1.0}  
    \centering
    \scriptsize
    \begin{tabular}{cc}
        \toprule
        Original & no text  \\
                 & Spatial bpp: $0.0019$, Text bpp: $0.0$bpp  \\
        \midrule

        \begin{subfigure}{0.45\textwidth}
            \centering
            \includegraphics[width=\linewidth]{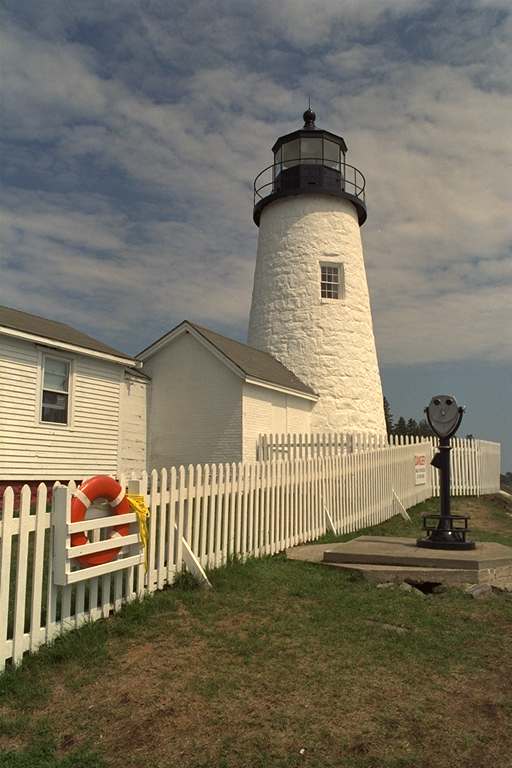}
        \end{subfigure}
        &
        \begin{subfigure}{0.45\textwidth}
            \centering
            \includegraphics[width=\linewidth]{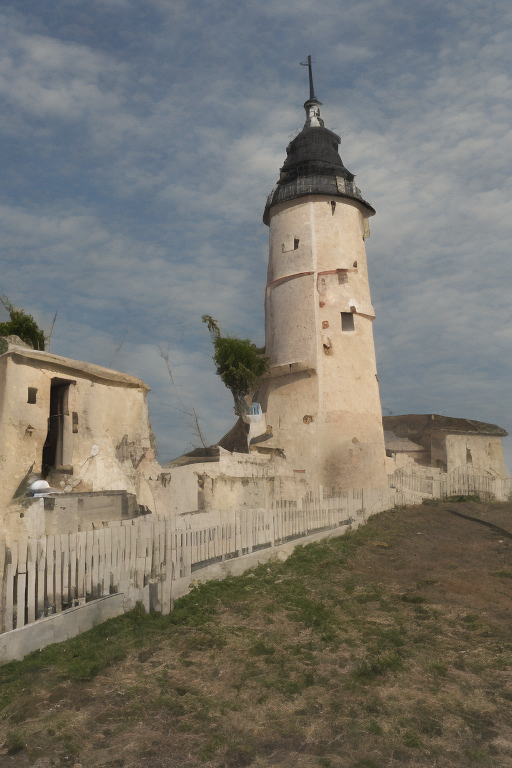}
        \end{subfigure}\\

        "a white fence with a lighthouse behind it" (BLIP 2) & "an old castle"  \\
        Spatial bpp: $0.0019$, Text bpp: $0.0010$bpp & Spatial bpp: $0.0019$, Text bpp: $0.0004$bpp \\
        \begin{subfigure}{0.45\textwidth}
            \centering
            \includegraphics[width=\linewidth]{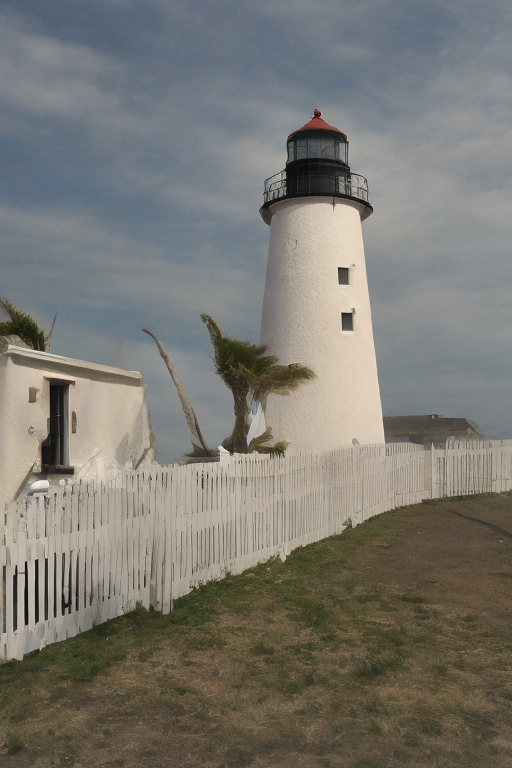}
        \end{subfigure}
        &
        \begin{subfigure}{0.45\textwidth}
            \centering
            \includegraphics[width=\linewidth]{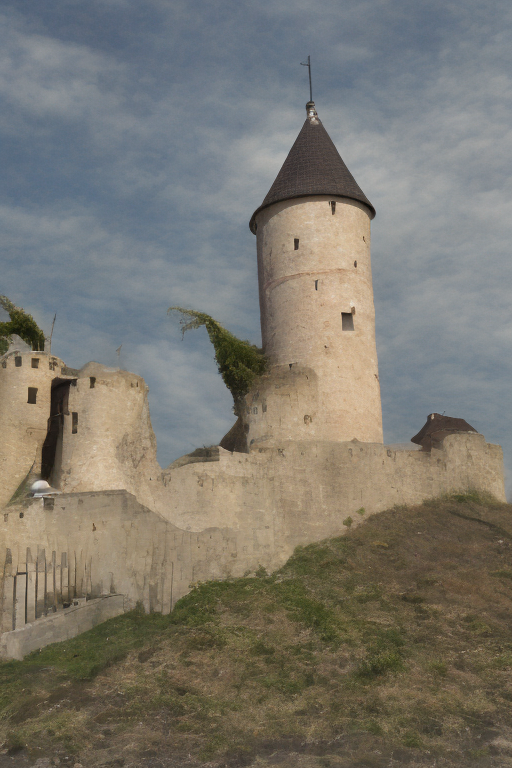}
        \end{subfigure}\\

        \bottomrule

    \end{tabular}

    \caption{Visual illustration of the impact of the global conditioning on the Kodak dataset (\texttt{kodim19}), with a spatial bit-rate of $0.0019$bpp. Samples are generated from the same initial Gaussian noise. Inspiration taken from~\cite[fig. 13]{careil2024towards}.}
    \label{fig:global_cond}
\end{figure*}

\textbf{Global conditioning.} In~\cref{fig:global_cond}, we analyze the impact of the global conditioning and show that PerCo (SD) offers similar internal characteristics.

\begin{figure*}[ht]
    \setlength{\tabcolsep}{1.0pt}  
    \renewcommand{\arraystretch}{1.0}  
    \centering
    \scriptsize
    \begin{tabular}{cc}
        \toprule
        PerCo & PerCo (SD)  \\
        ($0.003$bpp) & ($0.003$bpp) \\
        \midrule

        \begin{subfigure}{0.45\textwidth}
            \centering
            \includegraphics[width=\linewidth]{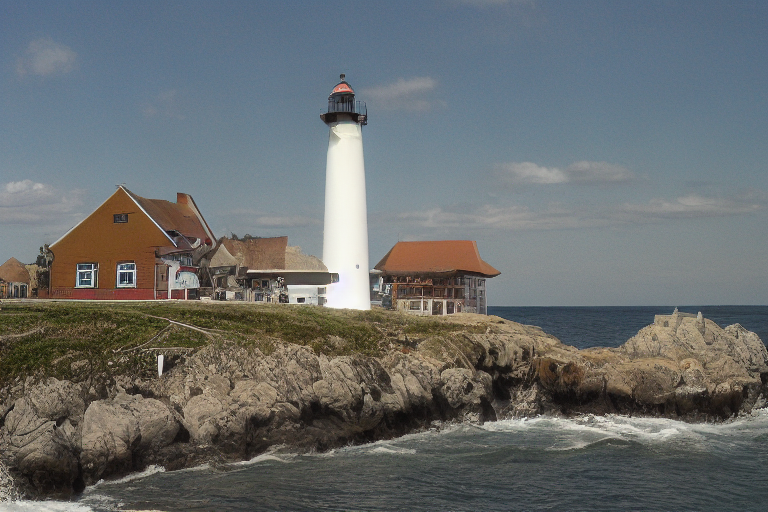}
        \end{subfigure}
        &
        \begin{subfigure}{0.45\textwidth}
            \centering
            \includegraphics[width=\linewidth]{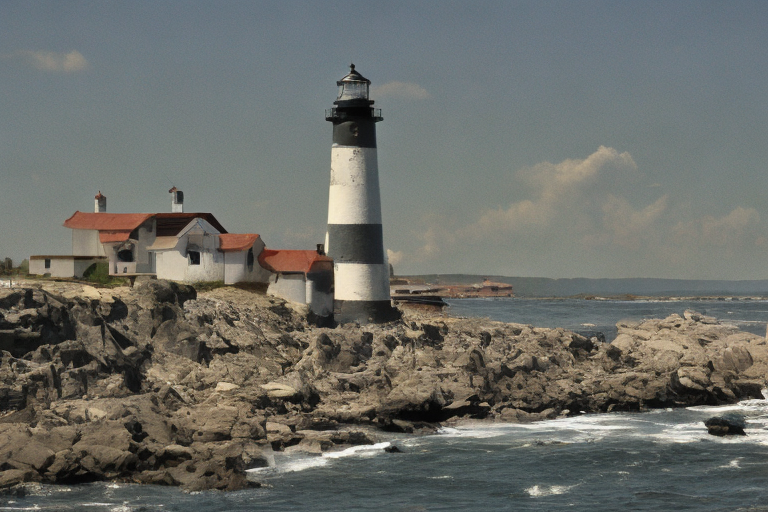}
        \end{subfigure}\\

        \begin{subfigure}{0.45\textwidth}
            \centering
            \includegraphics[width=\linewidth]{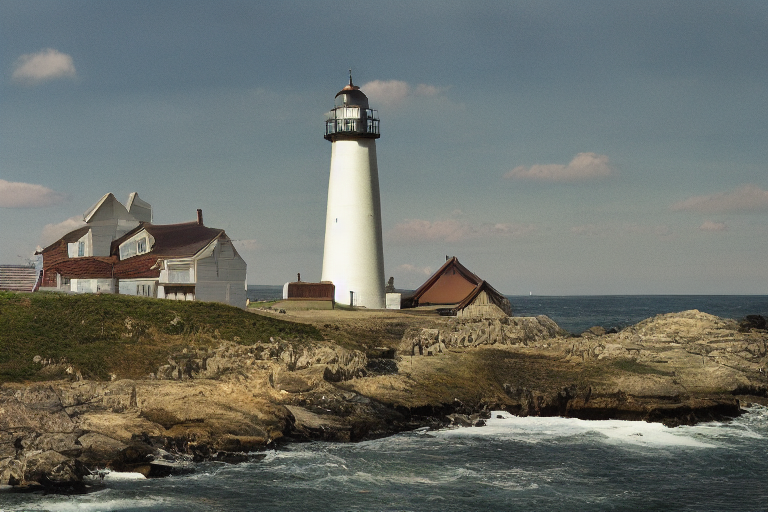}
        \end{subfigure}
        &
        \begin{subfigure}{0.45\textwidth}
            \centering
            \includegraphics[width=\linewidth]{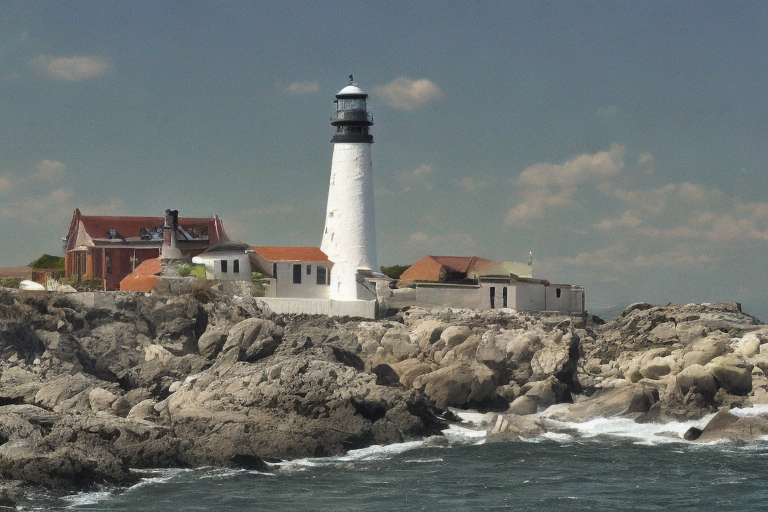}
        \end{subfigure}\\

        \begin{subfigure}{0.45\textwidth}
            \centering
            \includegraphics[width=\linewidth]{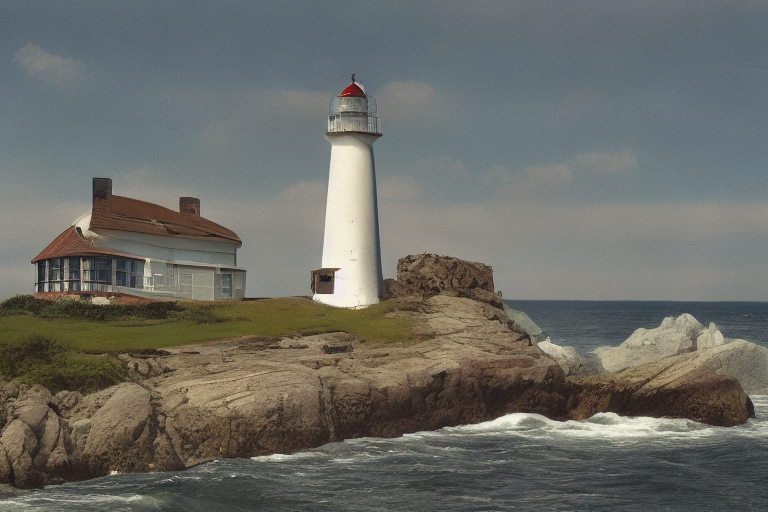}
        \end{subfigure}
        &
        \begin{subfigure}{0.45\textwidth}
            \centering
            \includegraphics[width=\linewidth]{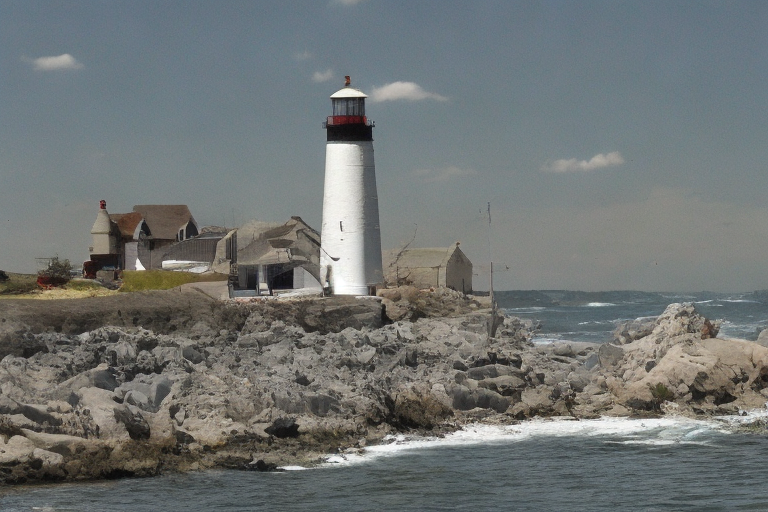}
        \end{subfigure}\\

        \begin{subfigure}{0.45\textwidth}
            \centering
            \includegraphics[width=\linewidth]{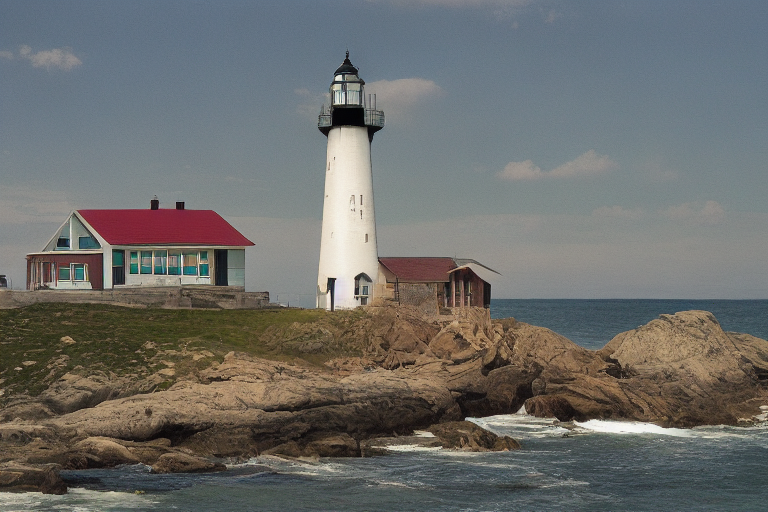}
        \end{subfigure}
        &
        \begin{subfigure}{0.45\textwidth}
            \centering
            \includegraphics[width=\linewidth]{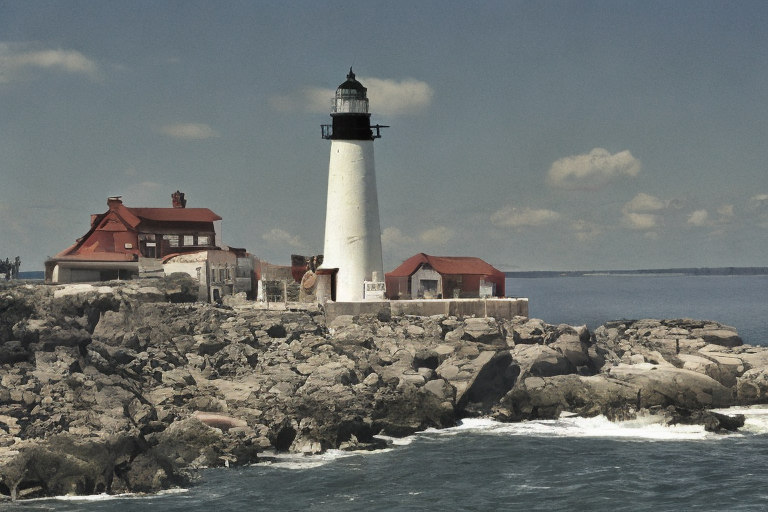}
        \end{subfigure}\\

        \bottomrule

    \end{tabular}

    \caption{Visual comparison of the sample diversity between PerCo~\cite{careil2024towards} and PerCo (SD)}
    \label{fig:diversity}
\end{figure*}

\textbf{Diversity.} In~\cref{fig:diversity}, we show a set of reconstructions from the same conditioning information that reflect the uncertainty about the original image source.

\begin{figure*}[ht]
    \setlength{\tabcolsep}{1.0pt}  
    \renewcommand{\arraystretch}{1.0}  
    \centering
    \scriptsize
    \begin{tabular}{cccc|c}
        \toprule
        $0.0036$bpp & $0.011$bpp & $0.0329$bpp & $0.1267$bpp & Original \\
        \midrule

        \multicolumn{5}{c}{Global conditioning: "a river runs through a rocky forest with a mountain in the background"} \\
        \begin{subfigure}{0.195\textwidth}
            \centering
            \includegraphics[width=\linewidth]{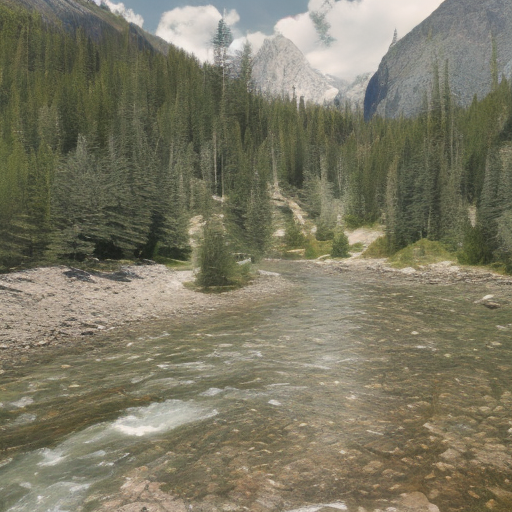}
        \end{subfigure}
        &
        \begin{subfigure}{0.195\textwidth}
            \centering
            \includegraphics[width=\linewidth]{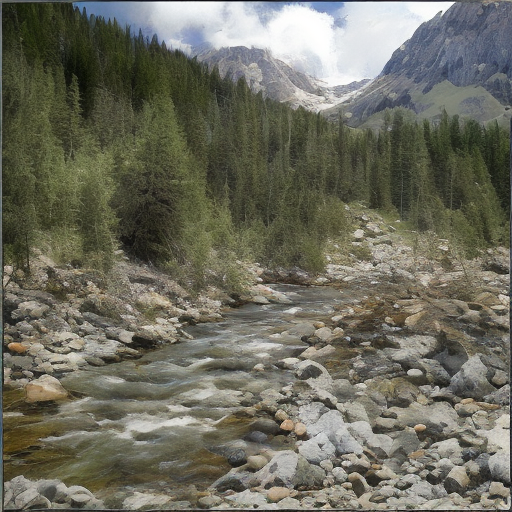}
        \end{subfigure}
        &
        \begin{subfigure}{0.195\textwidth}
            \centering
            \includegraphics[width=\linewidth]{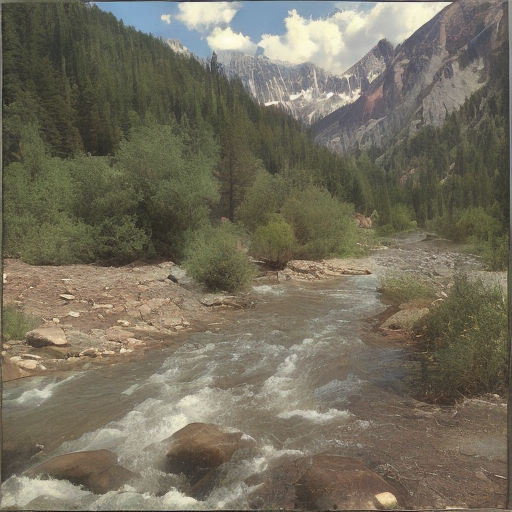}
        \end{subfigure}
        &
        \begin{subfigure}{0.195\textwidth}
            \centering
            \includegraphics[width=\linewidth]{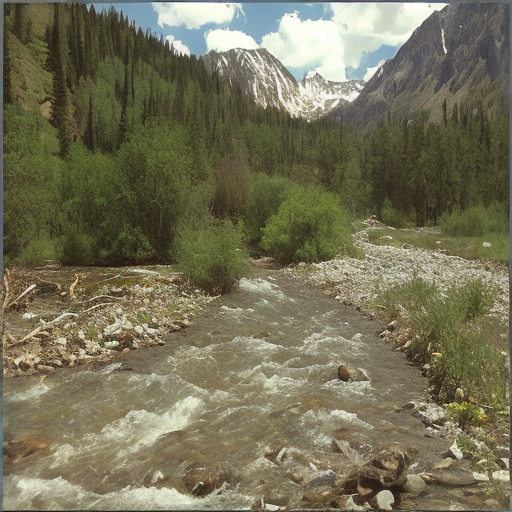}
        \end{subfigure}
        &
        \begin{subfigure}{0.195\textwidth}
            \centering
            \includegraphics[width=\linewidth]{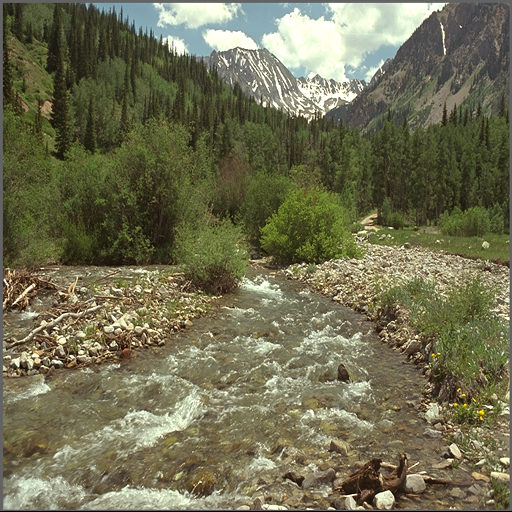}
        \end{subfigure} \\

        \multicolumn{5}{c}{Global conditioning: "a red barn with a pond in the background"} \\
        \begin{subfigure}{0.195\textwidth}
            \centering
            \includegraphics[width=\linewidth]{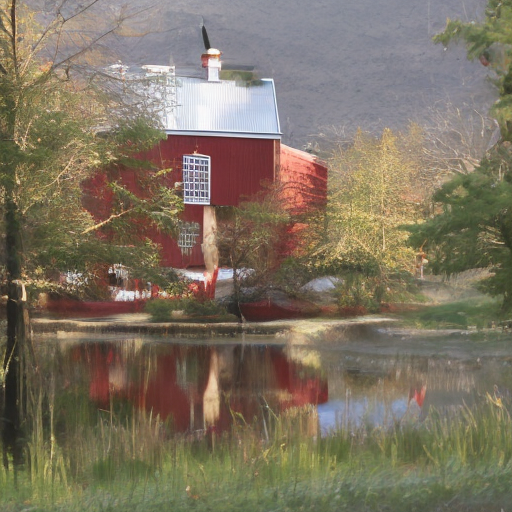}
        \end{subfigure}
        &
        \begin{subfigure}{0.195\textwidth}
            \centering
            \includegraphics[width=\linewidth]{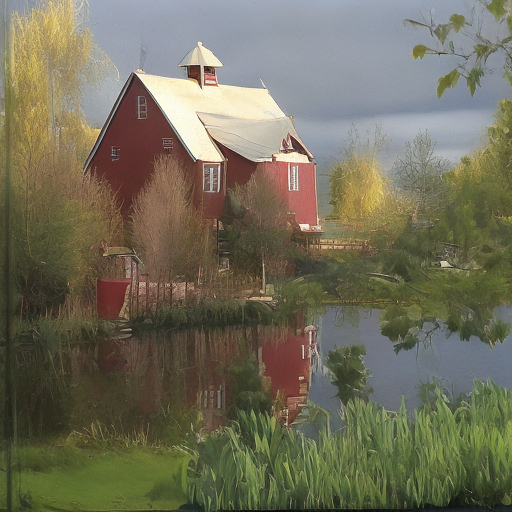}
        \end{subfigure}
        &
        \begin{subfigure}{0.195\textwidth}
            \centering
            \includegraphics[width=\linewidth]{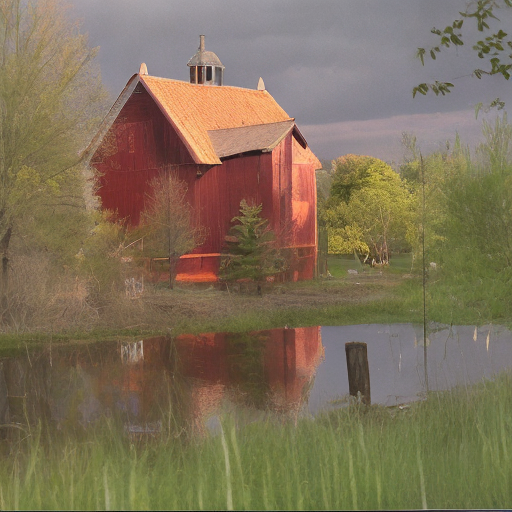}
        \end{subfigure}
        &
        \begin{subfigure}{0.195\textwidth}
            \centering
            \includegraphics[width=\linewidth]{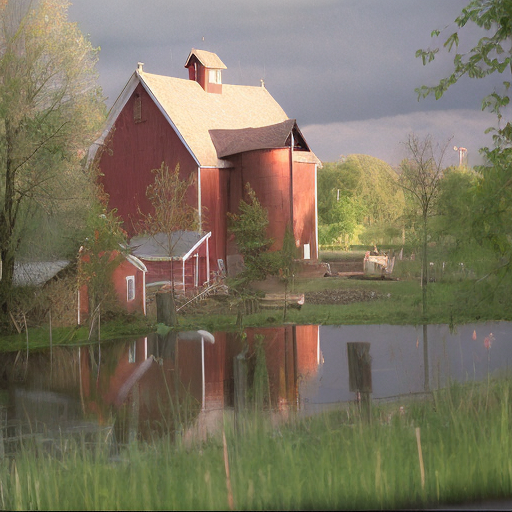}
        \end{subfigure}
        &
        \begin{subfigure}{0.195\textwidth}
            \centering
            \includegraphics[width=\linewidth]{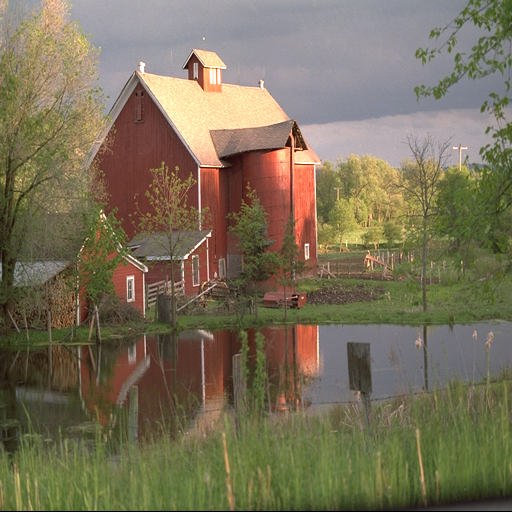}
        \end{subfigure} \\

        \multicolumn{5}{c}{Global conditioning: "a white fence with a lighthouse behind it"} \\
        \begin{subfigure}{0.195\textwidth}
            \centering
            \includegraphics[width=\linewidth]{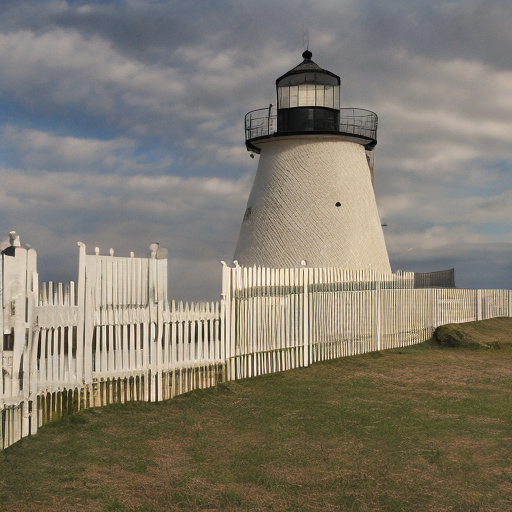}
        \end{subfigure}
        &
        \begin{subfigure}{0.195\textwidth}
            \centering
            \includegraphics[width=\linewidth]{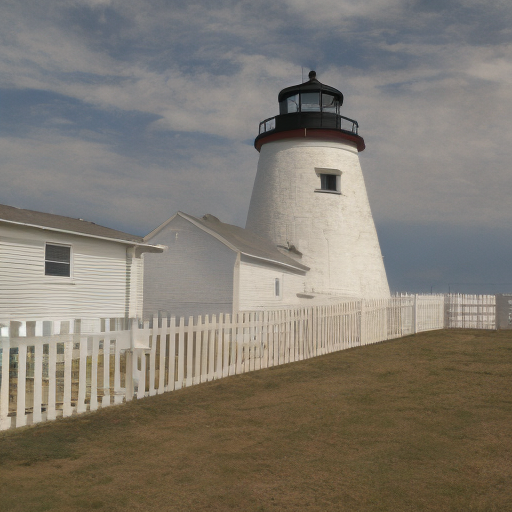}
        \end{subfigure}
        &
        \begin{subfigure}{0.195\textwidth}
            \centering
            \includegraphics[width=\linewidth]{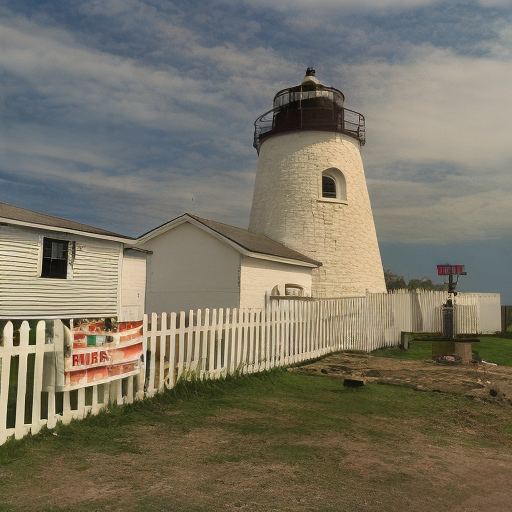}
        \end{subfigure}
        &
        \begin{subfigure}{0.195\textwidth}
            \centering
            \includegraphics[width=\linewidth]{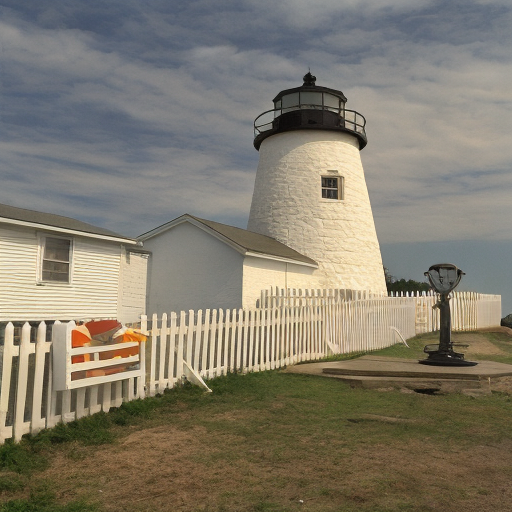}
        \end{subfigure}
        &
        \begin{subfigure}{0.195\textwidth}
            \centering
            \includegraphics[width=\linewidth]{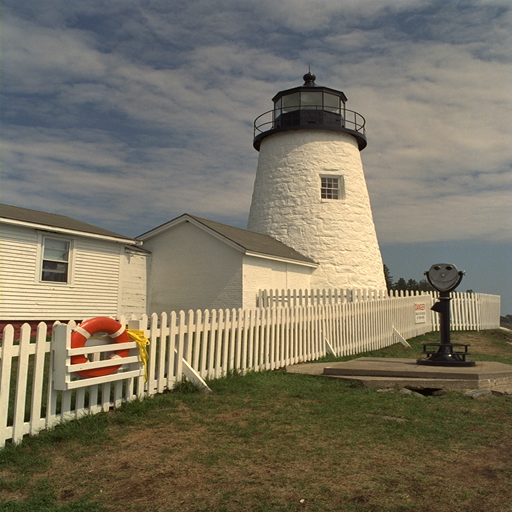}
        \end{subfigure} \\

        \multicolumn{5}{c}{Global conditioning: "a pink flower is in front of a window with blue shutters"} \\
        \begin{subfigure}{0.195\textwidth}
            \centering
            \includegraphics[width=\linewidth]{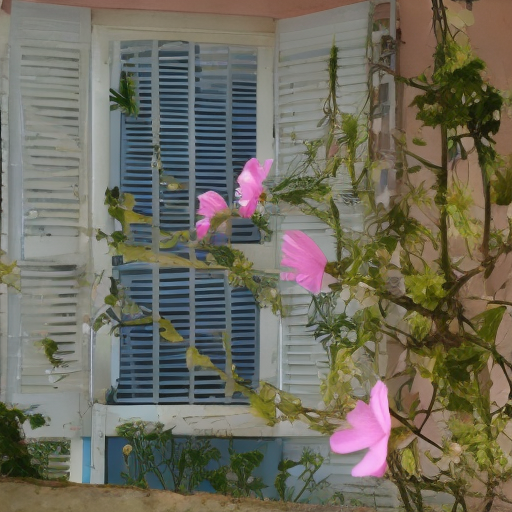}
        \end{subfigure}
        &
        \begin{subfigure}{0.195\textwidth}
            \centering
            \includegraphics[width=\linewidth]{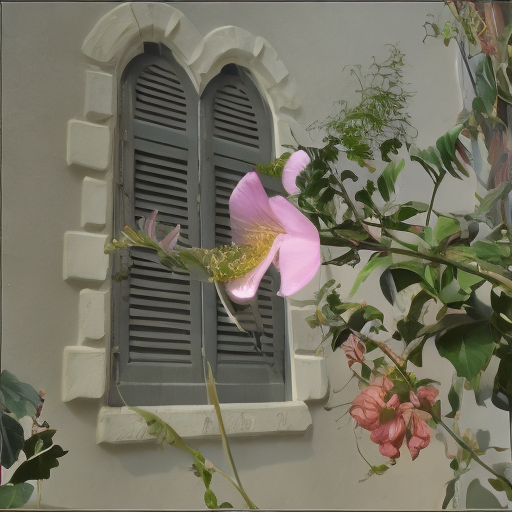}
        \end{subfigure}
        &
        \begin{subfigure}{0.195\textwidth}
            \centering
            \includegraphics[width=\linewidth]{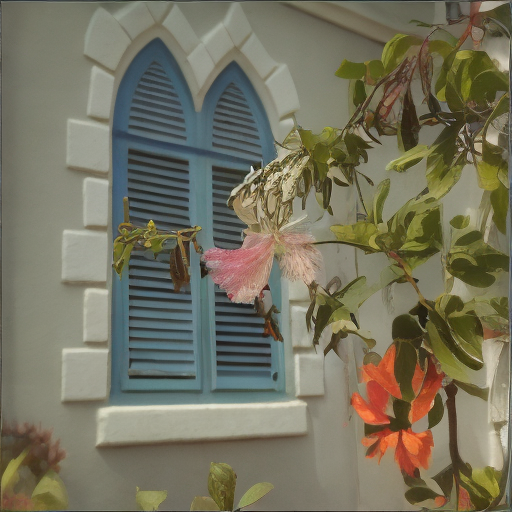}
        \end{subfigure}
        &
        \begin{subfigure}{0.195\textwidth}
            \centering
            \includegraphics[width=\linewidth]{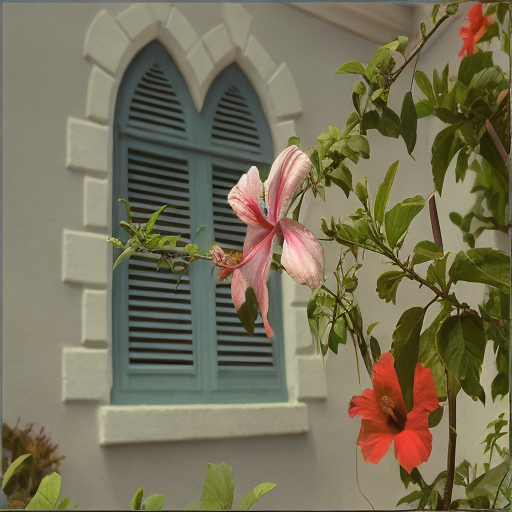}
        \end{subfigure}
        &
        \begin{subfigure}{0.195\textwidth}
            \centering
            \includegraphics[width=\linewidth]{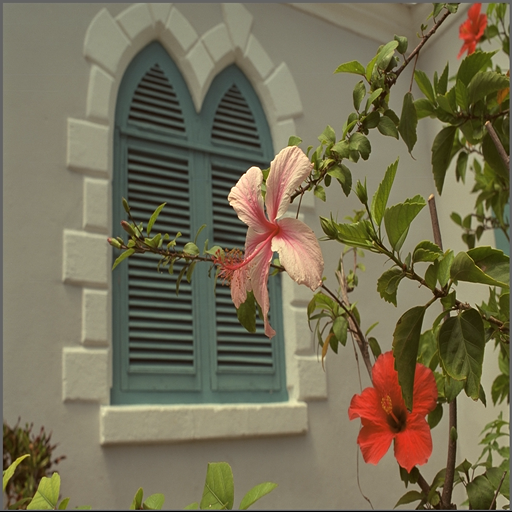}
        \end{subfigure} \\

        \multicolumn{5}{c}{Global conditioning: "a house with a balcony and a painting on the wall"} \\
        \begin{subfigure}{0.195\textwidth}
            \centering
            \includegraphics[width=\linewidth]{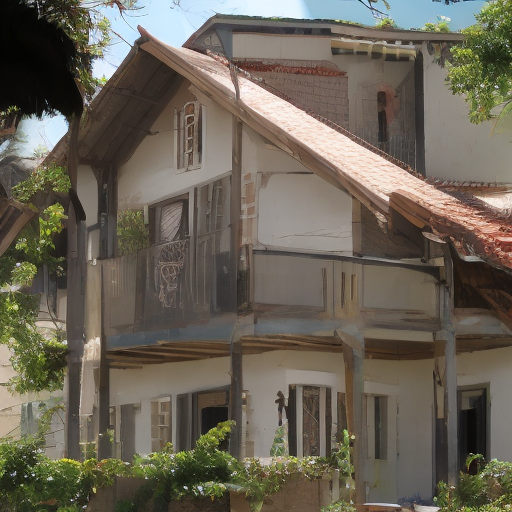}
        \end{subfigure}
        &
        \begin{subfigure}{0.195\textwidth}
            \centering
            \includegraphics[width=\linewidth]{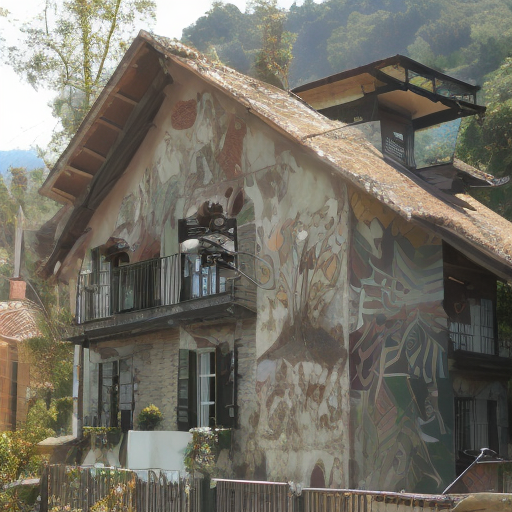}
        \end{subfigure}
        &
        \begin{subfigure}{0.195\textwidth}
            \centering
            \includegraphics[width=\linewidth]{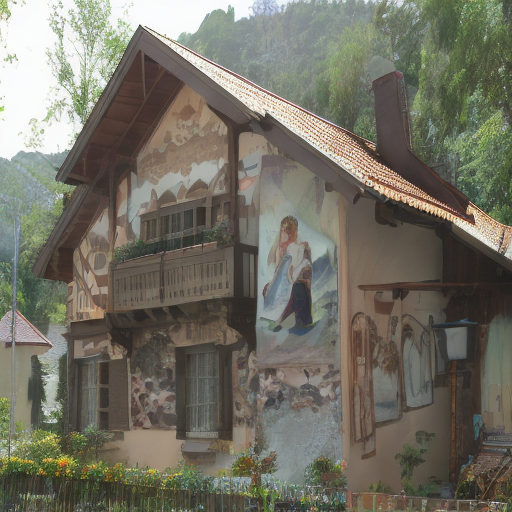}
        \end{subfigure}
        &
        \begin{subfigure}{0.195\textwidth}
            \centering
            \includegraphics[width=\linewidth]{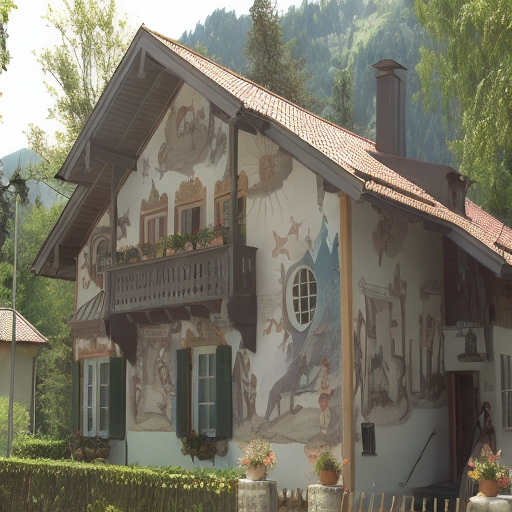}
        \end{subfigure}
        &
        \begin{subfigure}{0.195\textwidth}
            \centering
            \includegraphics[width=\linewidth]{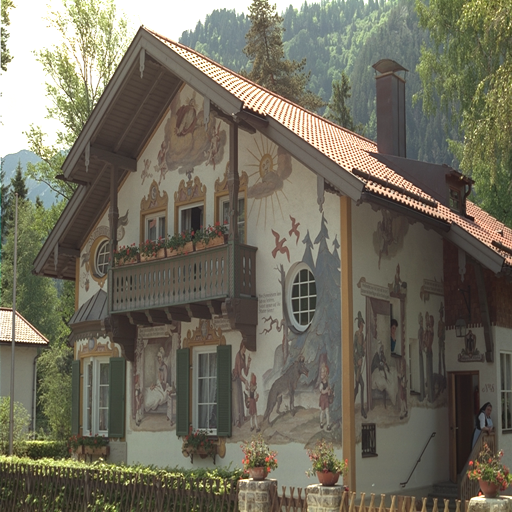}
        \end{subfigure} \\

        \multicolumn{5}{c}{Global conditioning: "a group of people on sailboats in the water"} \\
        \begin{subfigure}{0.195\textwidth}
            \centering
            \includegraphics[width=\linewidth]{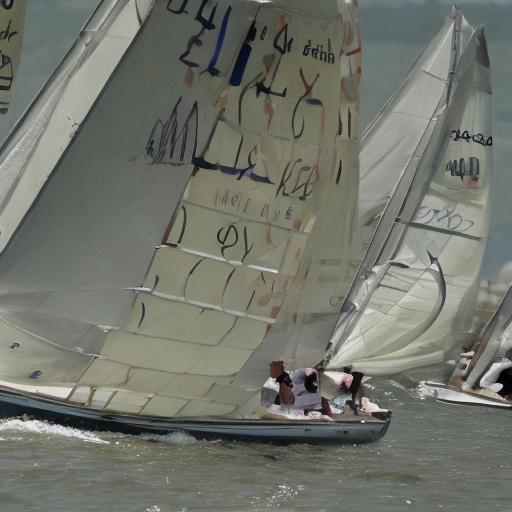}
        \end{subfigure}
        &
        \begin{subfigure}{0.195\textwidth}
            \centering
            \includegraphics[width=\linewidth]{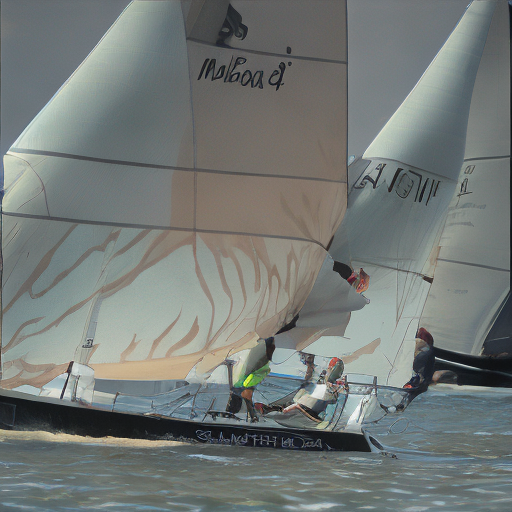}
        \end{subfigure}
        &
        \begin{subfigure}{0.195\textwidth}
            \centering
            \includegraphics[width=\linewidth]{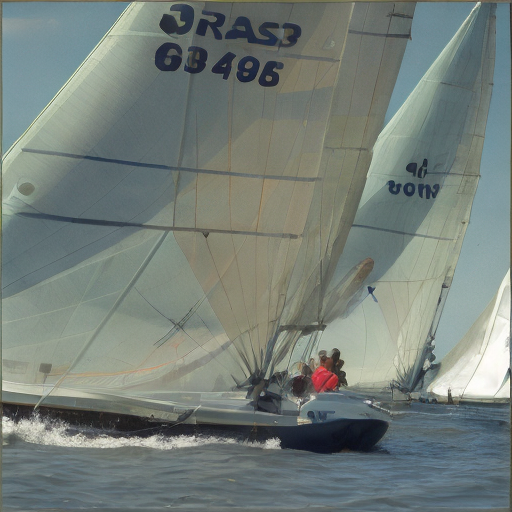}
        \end{subfigure}
        &
        \begin{subfigure}{0.195\textwidth}
            \centering
            \includegraphics[width=\linewidth]{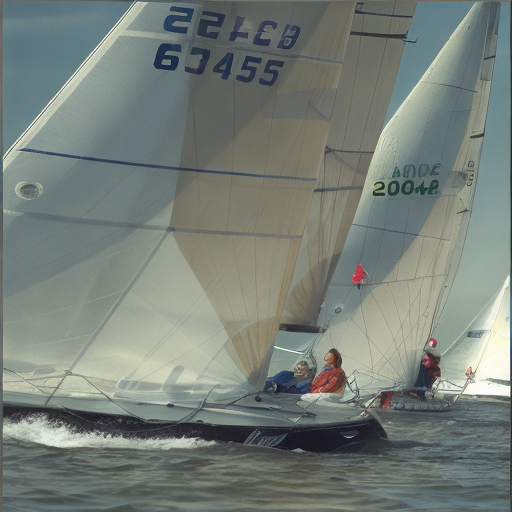}
        \end{subfigure}
        &
        \begin{subfigure}{0.195\textwidth}
            \centering
            \includegraphics[width=\linewidth]{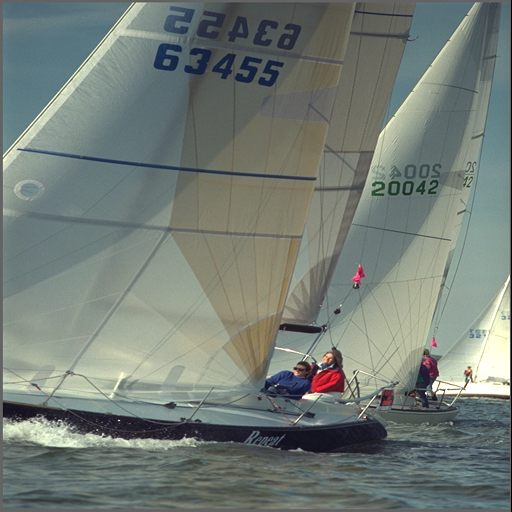}
        \end{subfigure} \\

        \bottomrule

    \end{tabular}

    \caption{Visual comparison of PerCo (SD) across various bit-rates on the Kodak dataset}
    \label{fig:vis_bit_rate_interpolation}
\end{figure*}

\textbf{Reconstructions across various bit rates.} In~\cref{fig:vis_bit_rate_interpolation}, we visualize reconstructions with increasing access to local conditioning information.

\begin{figure*}[ht]
    \setlength{\tabcolsep}{1.0pt}  
    \renewcommand{\arraystretch}{1.0}  
    \centering
    \scriptsize
    \begin{tabular}{cccc}
        CFG $0.0$, steps $5$ & CFG $1.0$, steps $5$ & CFG $3.0$, steps $5$ & CFG $7.5$, steps $5$ \\
        \begin{subfigure}{0.245\textwidth}
            \centering
            \includegraphics[width=\linewidth]{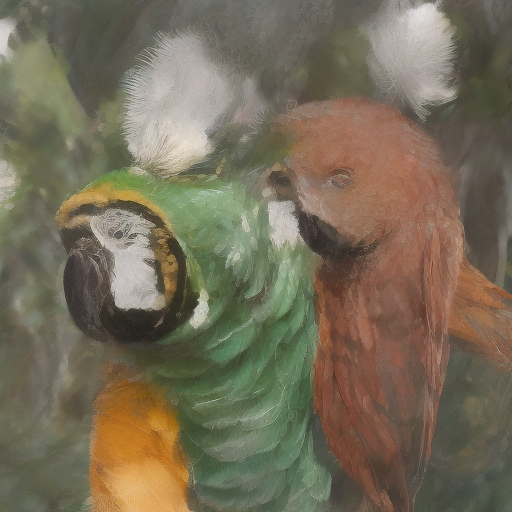}
        \end{subfigure}
        &
        \begin{subfigure}{0.245\textwidth}
            \centering
            \includegraphics[width=\linewidth]{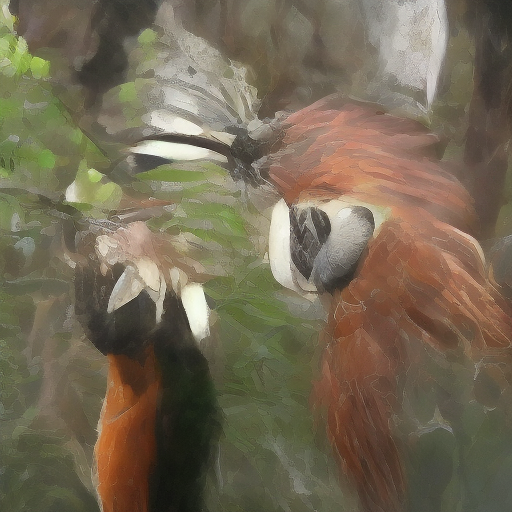}
        \end{subfigure}
        &
        \begin{subfigure}{0.245\textwidth}
            \centering
            \includegraphics[width=\linewidth]{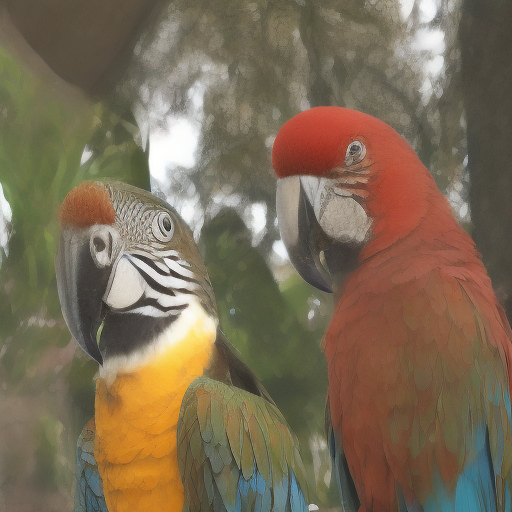}
        \end{subfigure}
        &
        \begin{subfigure}{0.245\textwidth}
            \centering
            \includegraphics[width=\linewidth]{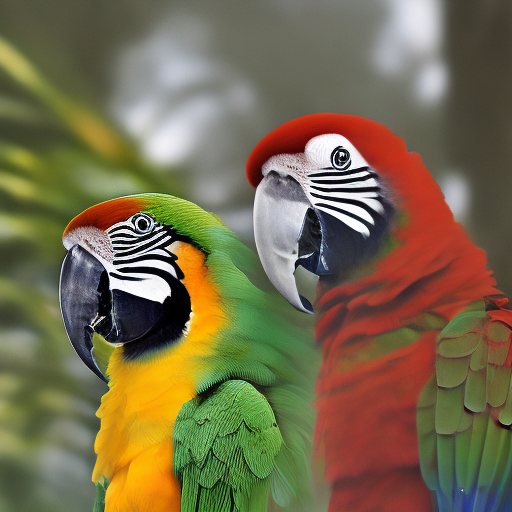}
        \end{subfigure} \\

        CFG $0.0$, steps $20$ & CFG $1.0$, steps $20$ & CFG $3.0$, steps $20$ & CFG $7.5$, steps $20$ \\
        \begin{subfigure}{0.245\textwidth}
            \centering
            \includegraphics[width=\linewidth]{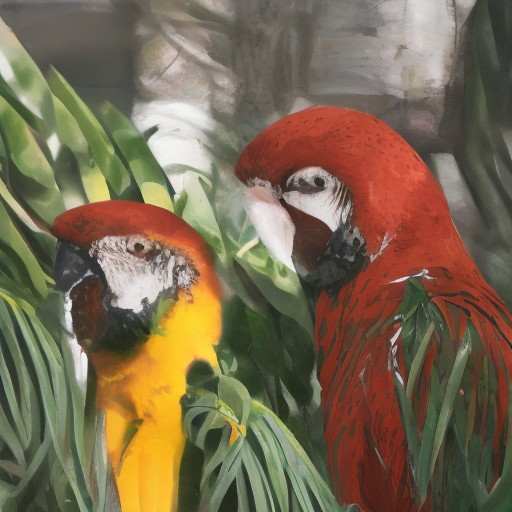}
        \end{subfigure}
        &
        \begin{subfigure}{0.245\textwidth}
            \centering
            \includegraphics[width=\linewidth]{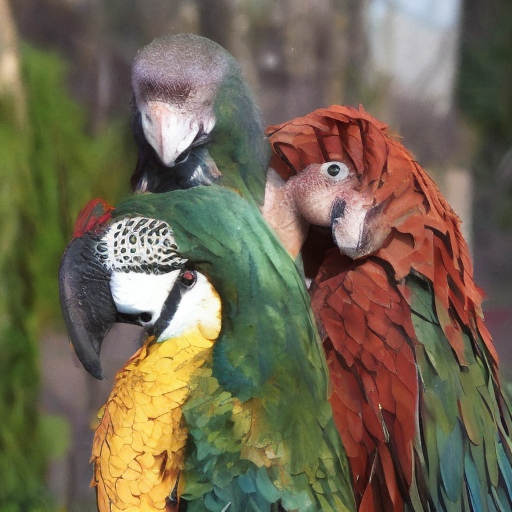}
        \end{subfigure}
        &
        \begin{subfigure}{0.245\textwidth}
            \centering
            \includegraphics[width=\linewidth]{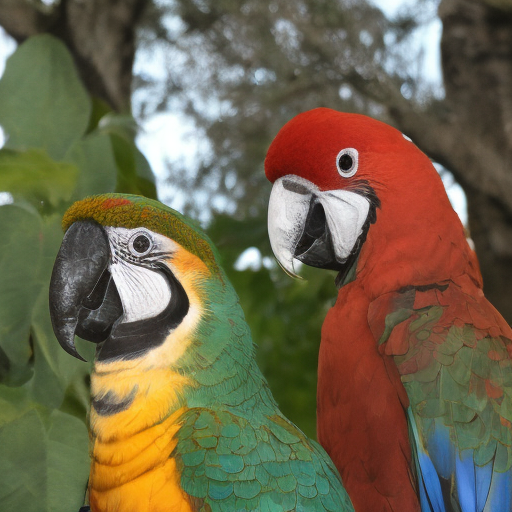}
        \end{subfigure}
        &
        \begin{subfigure}{0.245\textwidth}
            \centering
            \includegraphics[width=\linewidth]{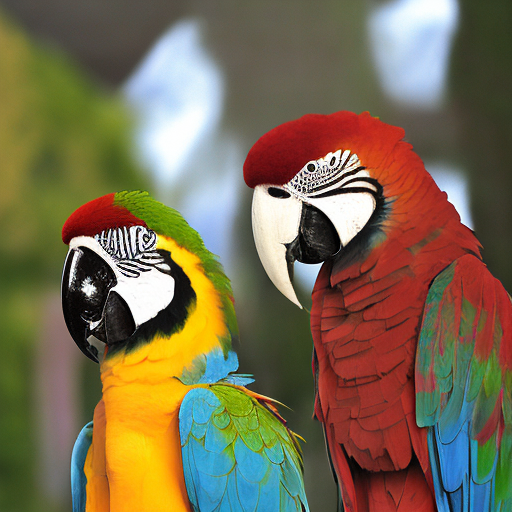}
        \end{subfigure} \\
        
        CFG $0.0$, steps $50$ & CFG $1.0$, steps $50$ & CFG $3.0$, steps $50$ & CFG $7.5$, steps $50$ \\
        \begin{subfigure}{0.245\textwidth}
            \centering
            \includegraphics[width=\linewidth]{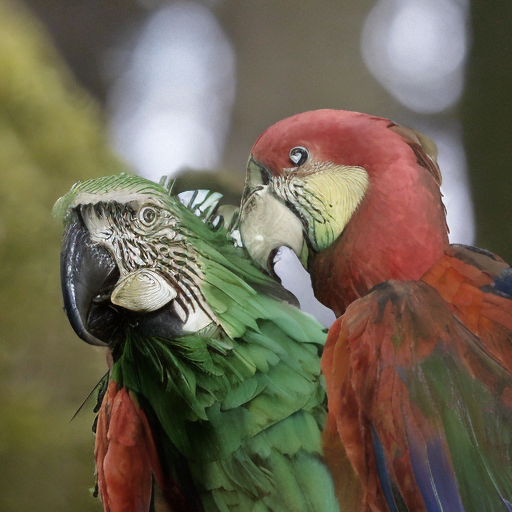}
        \end{subfigure}
        &
        \begin{subfigure}{0.245\textwidth}
            \centering
            \includegraphics[width=\linewidth]{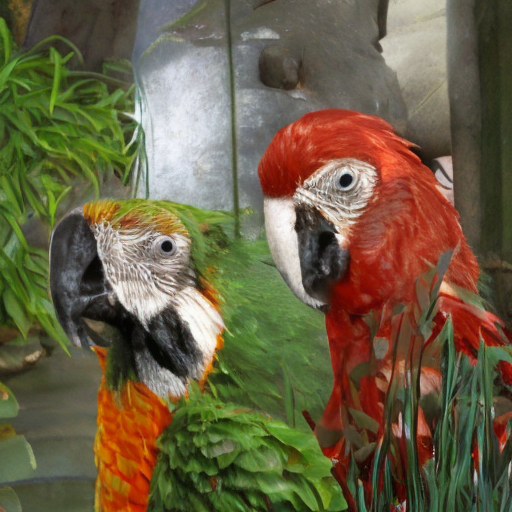}
        \end{subfigure}
        &
        \begin{subfigure}{0.245\textwidth}
            \centering
            \includegraphics[width=\linewidth]{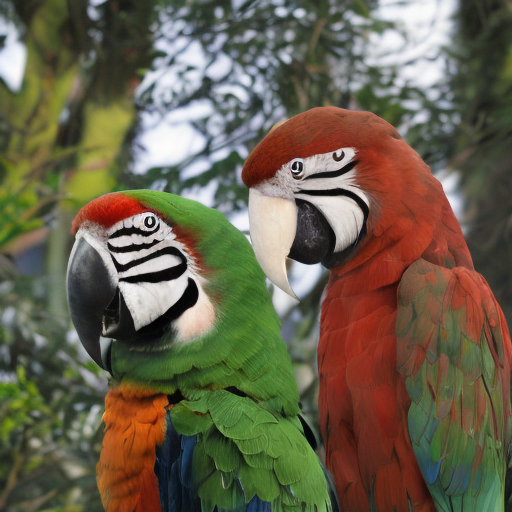}
        \end{subfigure}
        &
        \begin{subfigure}{0.245\textwidth}
            \centering
            \includegraphics[width=\linewidth]{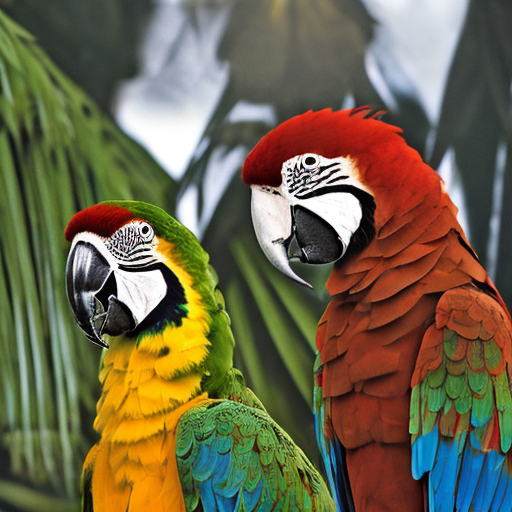}
        \end{subfigure} \\

        CFG $0.0$, steps $100$ & CFG $1.0$, steps $100$ & CFG $3.0$, steps $100$ & CFG $7.5$, steps $100$ \\
        \begin{subfigure}{0.245\textwidth}
            \centering
            \includegraphics[width=\linewidth]{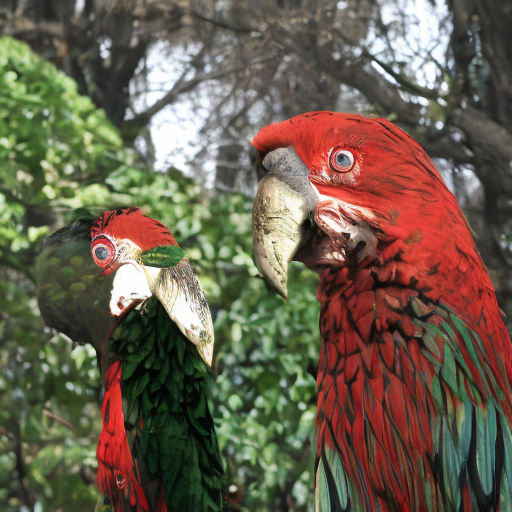}
        \end{subfigure}
        &
        \begin{subfigure}{0.245\textwidth}
            \centering
            \includegraphics[width=\linewidth]{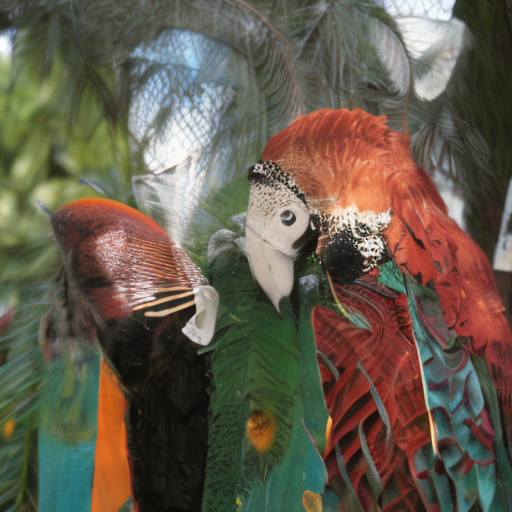}
        \end{subfigure}
        &
        \begin{subfigure}{0.245\textwidth}
            \centering
            \includegraphics[width=\linewidth]{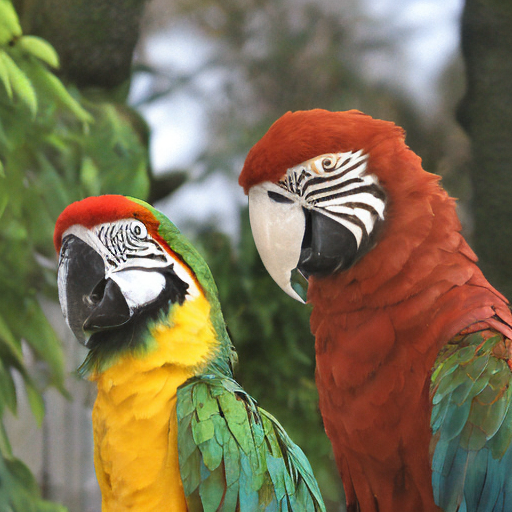}
        \end{subfigure}
        &
        \begin{subfigure}{0.245\textwidth}
            \centering
            \includegraphics[width=\linewidth]{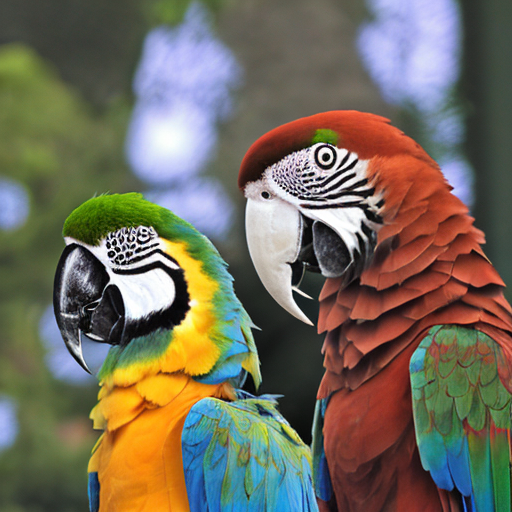}
        \end{subfigure} \\

    \end{tabular}

    \caption{Visual comparison of PerCo (SD) on the Kodak dataset (\texttt{kodim23}), illustrating the impact of classifier-free guidance~\cite{ho2021classifierfree} (CFG) and varying numbers of sampling steps at $0.0036$bpp.}
    \label{fig:cfg_vs_num_steps_1}
\end{figure*}

\begin{figure*}[ht]
    \setlength{\tabcolsep}{1.0pt}  
    \renewcommand{\arraystretch}{1.0}  
    \centering
    \scriptsize
    \begin{tabular}{cccc}
        CFG $0.0$, steps $5$ & CFG $1.0$, steps $5$ & CFG $3.0$, steps $5$ & CFG $7.5$, steps $5$ \\
        \begin{subfigure}{0.245\textwidth}
            \centering
            \includegraphics[width=\linewidth]{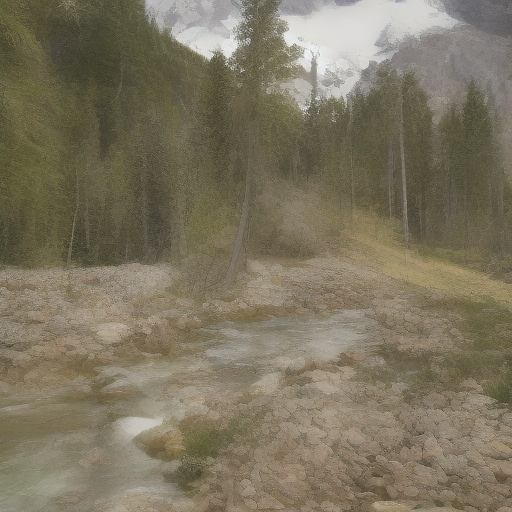}
        \end{subfigure}
        &
        \begin{subfigure}{0.245\textwidth}
            \centering
            \includegraphics[width=\linewidth]{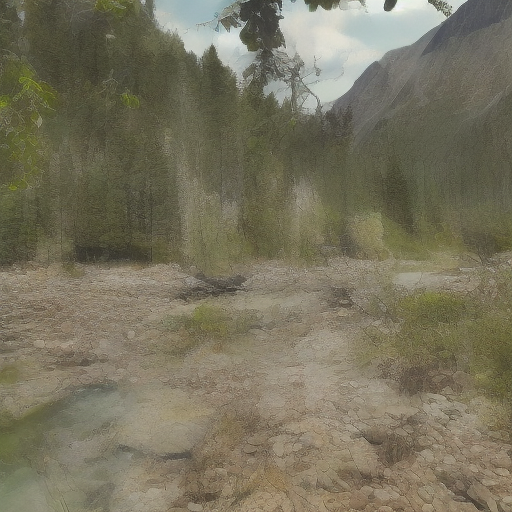}
        \end{subfigure}
        &
        \begin{subfigure}{0.245\textwidth}
            \centering
            \includegraphics[width=\linewidth]{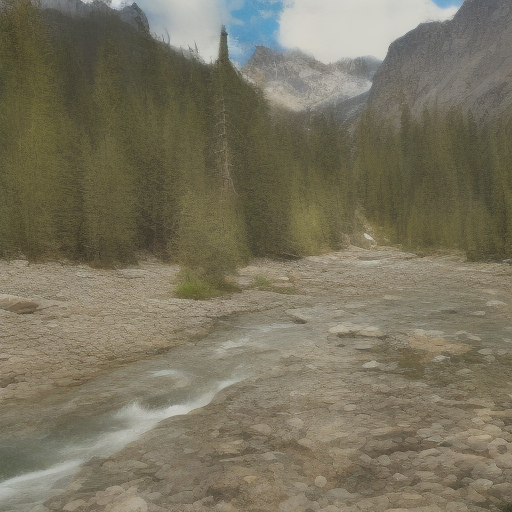}
        \end{subfigure}
        &
        \begin{subfigure}{0.245\textwidth}
            \centering
            \includegraphics[width=\linewidth]{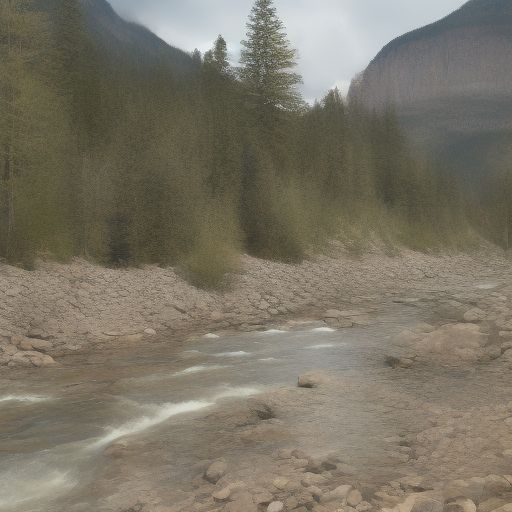}
        \end{subfigure} \\

        CFG $0.0$, steps $20$ & CFG $1.0$, steps $20$ & CFG $3.0$, steps $20$ & CFG $7.5$, steps $20$ \\
        \begin{subfigure}{0.245\textwidth}
            \centering
            \includegraphics[width=\linewidth]{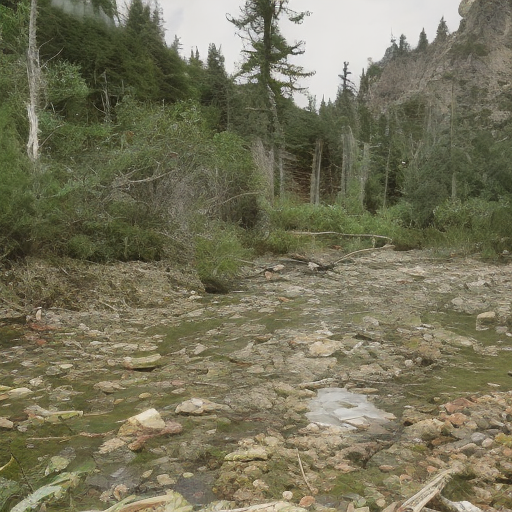}
        \end{subfigure}
        &
        \begin{subfigure}{0.245\textwidth}
            \centering
            \includegraphics[width=\linewidth]{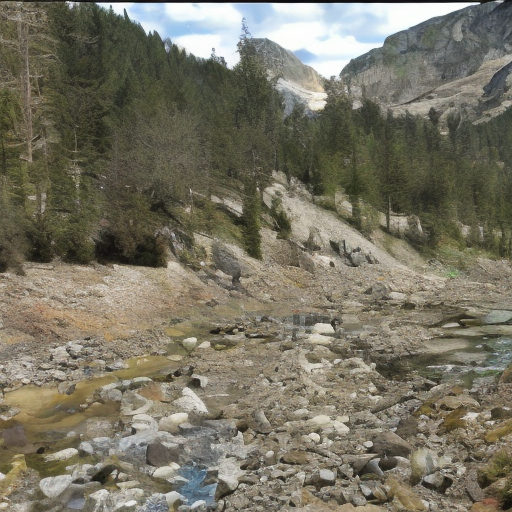}
        \end{subfigure}
        &
        \begin{subfigure}{0.245\textwidth}
            \centering
            \includegraphics[width=\linewidth]{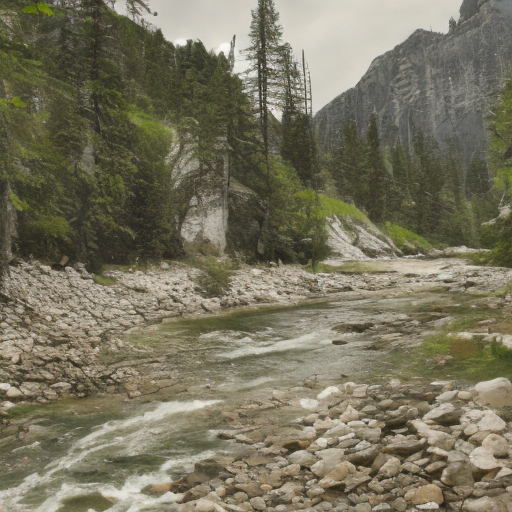}
        \end{subfigure}
        &
        \begin{subfigure}{0.245\textwidth}
            \centering
            \includegraphics[width=\linewidth]{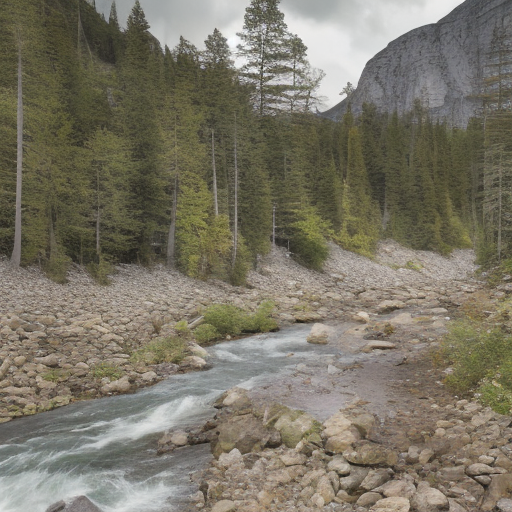}
        \end{subfigure} \\
        
        CFG $0.0$, steps $50$ & CFG $1.0$, steps $50$ & CFG $3.0$, steps $50$ & CFG $7.5$, steps $50$ \\
        \begin{subfigure}{0.245\textwidth}
            \centering
            \includegraphics[width=\linewidth]{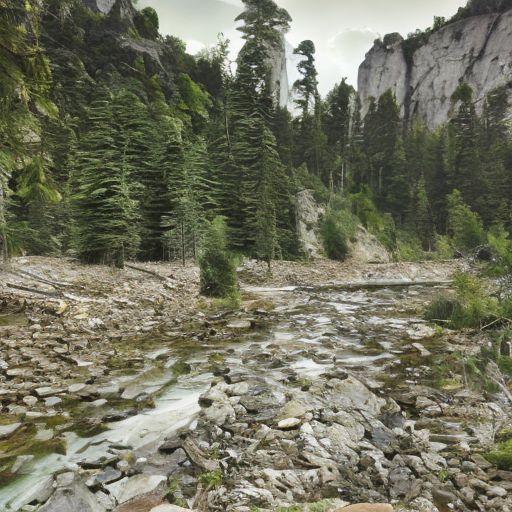}
        \end{subfigure}
        &
        \begin{subfigure}{0.245\textwidth}
            \centering
            \includegraphics[width=\linewidth]{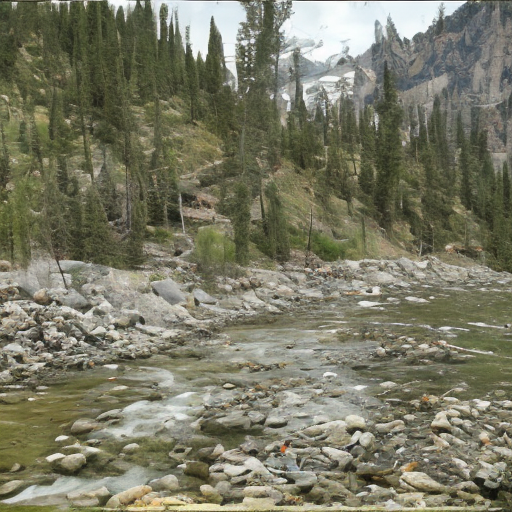}
        \end{subfigure}
        &
        \begin{subfigure}{0.245\textwidth}
            \centering
            \includegraphics[width=\linewidth]{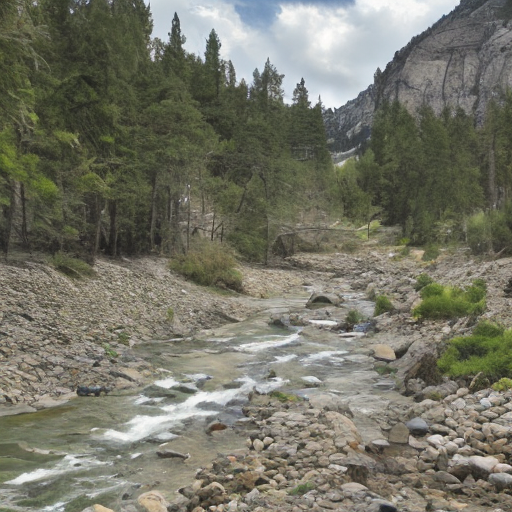}
        \end{subfigure}
        &
        \begin{subfigure}{0.245\textwidth}
            \centering
            \includegraphics[width=\linewidth]{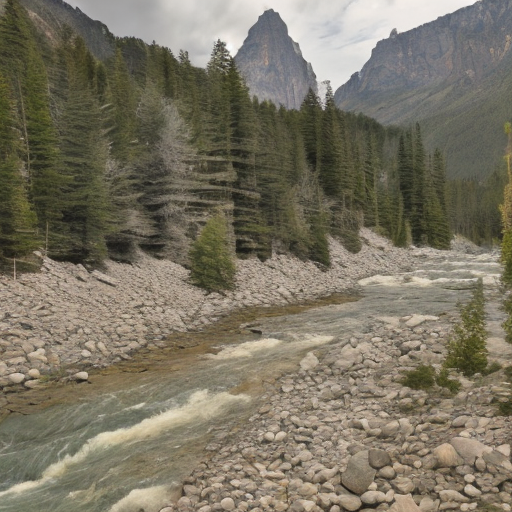}
        \end{subfigure} \\

        CFG $0.0$, steps $100$ & CFG $1.0$, steps $100$ & CFG $3.0$, steps $100$ & CFG $7.5$, steps $100$ \\
        \begin{subfigure}{0.245\textwidth}
            \centering
            \includegraphics[width=\linewidth]{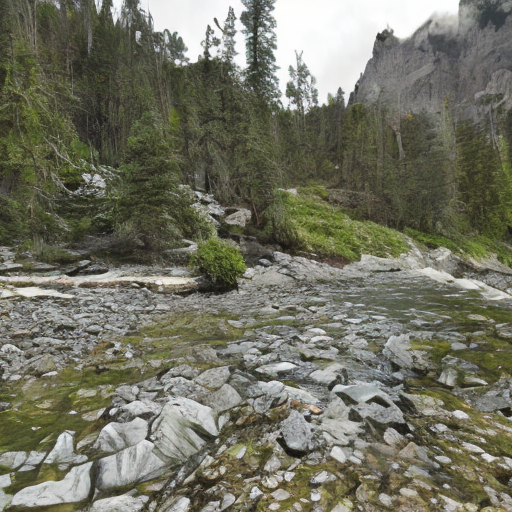}
        \end{subfigure}
        &
        \begin{subfigure}{0.245\textwidth}
            \centering
            \includegraphics[width=\linewidth]{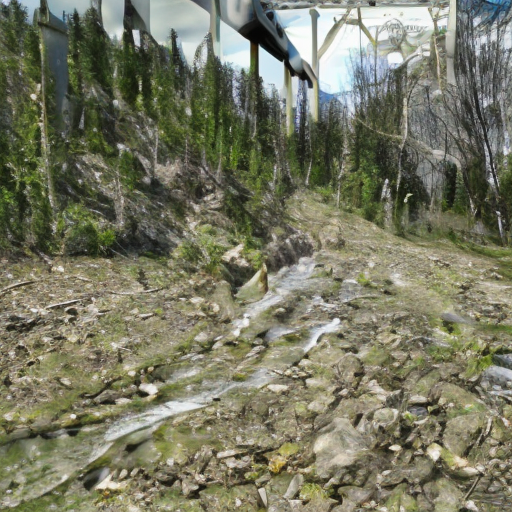}
        \end{subfigure}
        &
        \begin{subfigure}{0.245\textwidth}
            \centering
            \includegraphics[width=\linewidth]{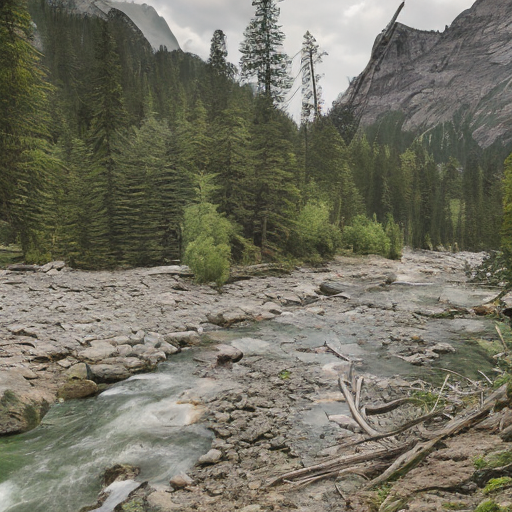}
        \end{subfigure}
        &
        \begin{subfigure}{0.245\textwidth}
            \centering
            \includegraphics[width=\linewidth]{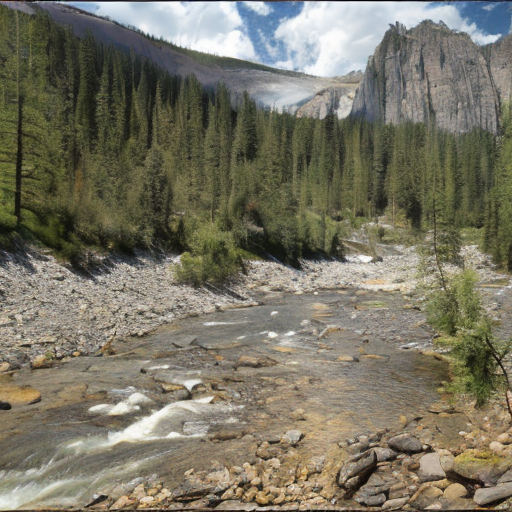}
        \end{subfigure} \\

    \end{tabular}

    \caption{Visual comparison of PerCo (SD) on the Kodak dataset (\texttt{kodim13}), illustrating the impact of classifier-free guidance~\cite{ho2021classifierfree} (CFG) and varying numbers of sampling steps at $0.0036$bpp.}
    \label{fig:cfg_vs_num_steps_2}
\end{figure*}

\textbf{Classifier-free guidance vs. steps.} In~\cref{fig:cfg_vs_num_steps_1} and~\cref{fig:cfg_vs_num_steps_2}, we show the interaction between various classifier-free guidance scales and number of sampling steps.

\begin{figure*}[ht]
    \setlength{\tabcolsep}{1.0pt}  
    \renewcommand{\arraystretch}{1.0}  
    \centering
    \scriptsize
    \begin{tabular}{cc|cc}

        Original & GT label & $0.00363$bpp & Predicted label \\
        \begin{subfigure}{0.245\textwidth}
            \centering
            \includegraphics[width=\linewidth]{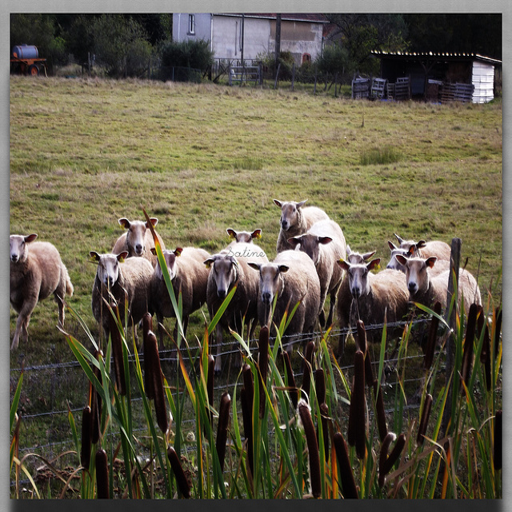}
        \end{subfigure}
        &
        \begin{subfigure}{0.245\textwidth}
            \centering
            \includegraphics[width=\linewidth]{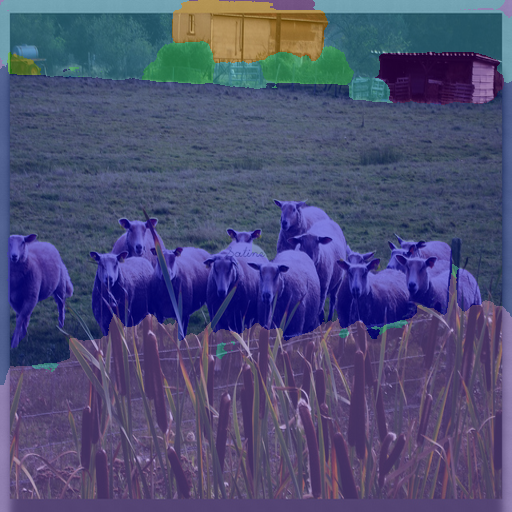}
        \end{subfigure}
        &
        \begin{subfigure}{0.245\textwidth}
            \centering
            \includegraphics[width=\linewidth]{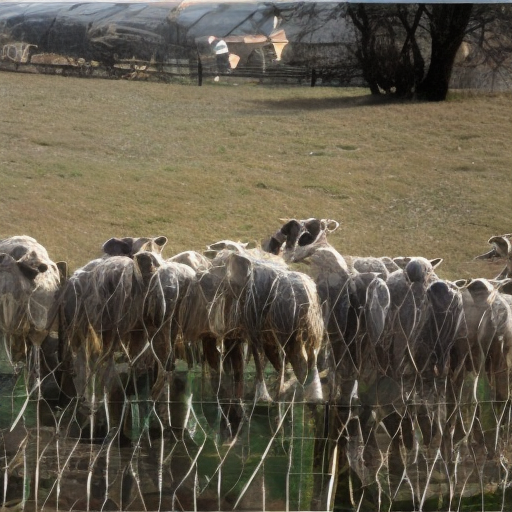}
        \end{subfigure}
        &
        \begin{subfigure}{0.245\textwidth}
            \centering
            \includegraphics[width=\linewidth]{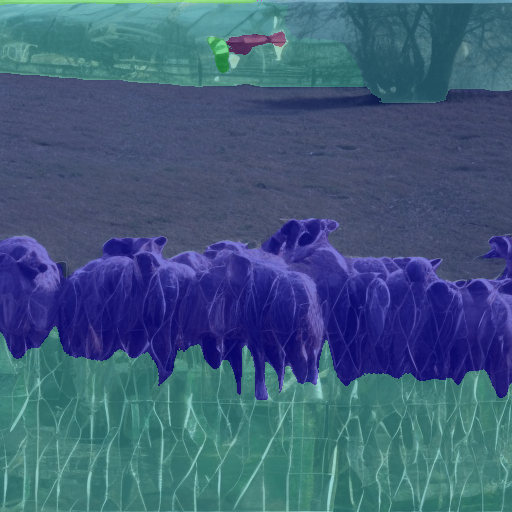}
        \end{subfigure} \\

        $0.00754$bpp & Predicted label & $0.01144$bpp & Predicted label \\
        \begin{subfigure}{0.245\textwidth}
            \centering
            \includegraphics[width=\linewidth]{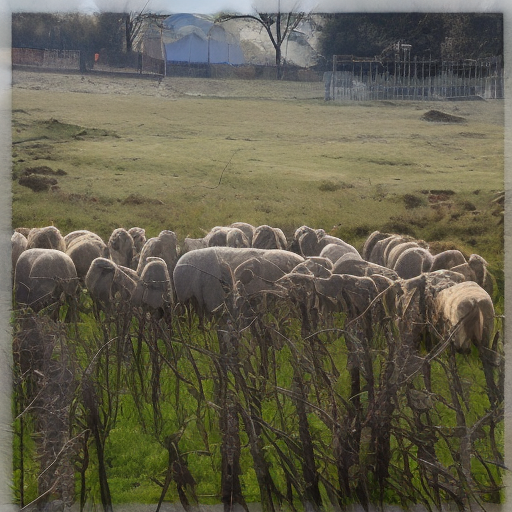}
        \end{subfigure}
        &
        \begin{subfigure}{0.245\textwidth}
            \centering
            \includegraphics[width=\linewidth]{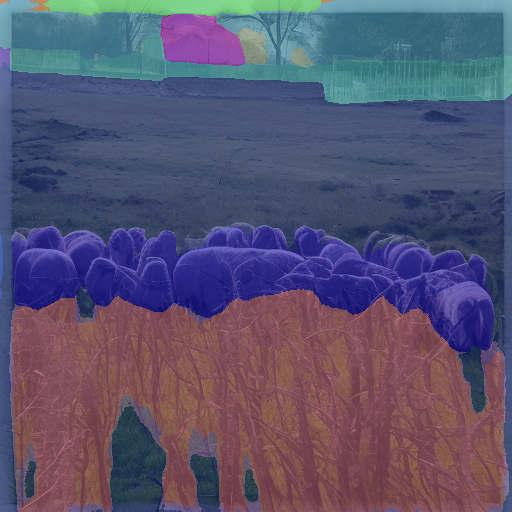}
        \end{subfigure}
        &
        \begin{subfigure}{0.245\textwidth}
            \centering
            \includegraphics[width=\linewidth]{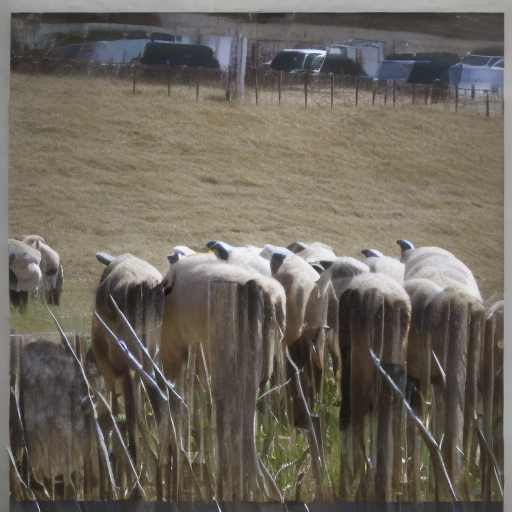}
        \end{subfigure}
        &
        \begin{subfigure}{0.245\textwidth}
            \centering
            \includegraphics[width=\linewidth]{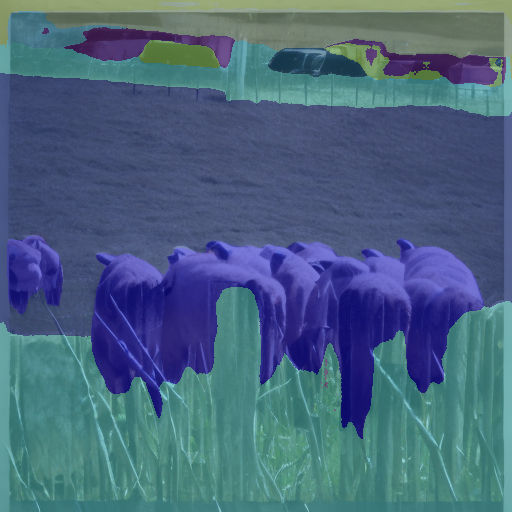}
        \end{subfigure} \\

        $0.03293$bpp & Predicted label & $0.05246$bpp & Predicted label \\
        \begin{subfigure}{0.245\textwidth}
            \centering
            \includegraphics[width=\linewidth]{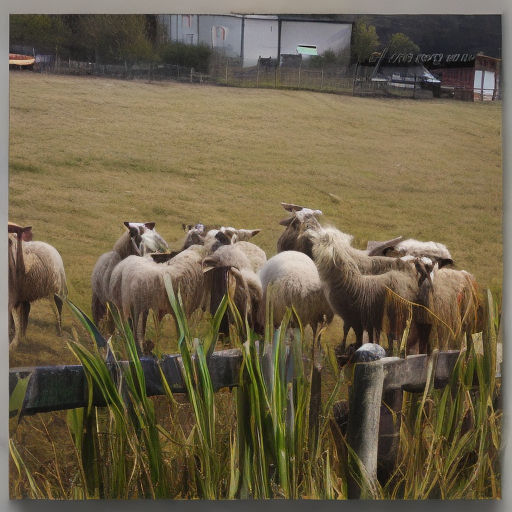}
        \end{subfigure}
        &
        \begin{subfigure}{0.245\textwidth}
            \centering
            \includegraphics[width=\linewidth]{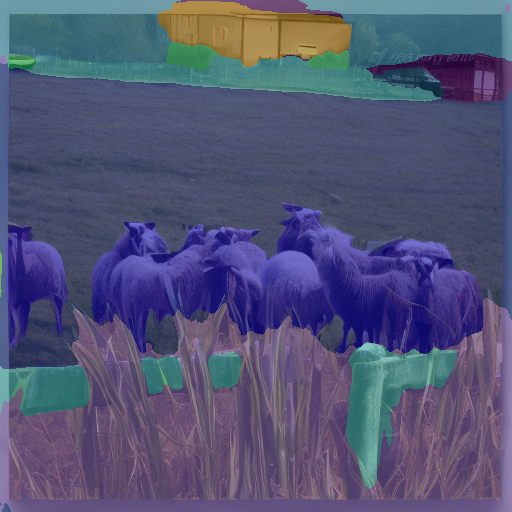}
        \end{subfigure}
        &
        \begin{subfigure}{0.245\textwidth}
            \centering
            \includegraphics[width=\linewidth]{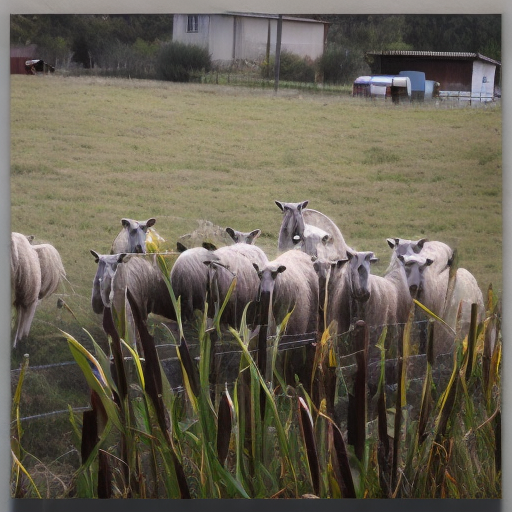}
        \end{subfigure}
        &
        \begin{subfigure}{0.245\textwidth}
            \centering
            \includegraphics[width=\linewidth]{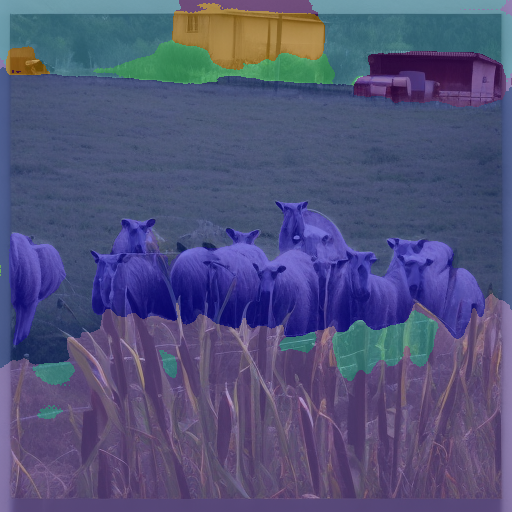}
        \end{subfigure} \\

        $0.09543$bpp & Predicted label & $0.12668$bpp & Predicted label \\
        \begin{subfigure}{0.245\textwidth}
            \centering
            \includegraphics[width=\linewidth]{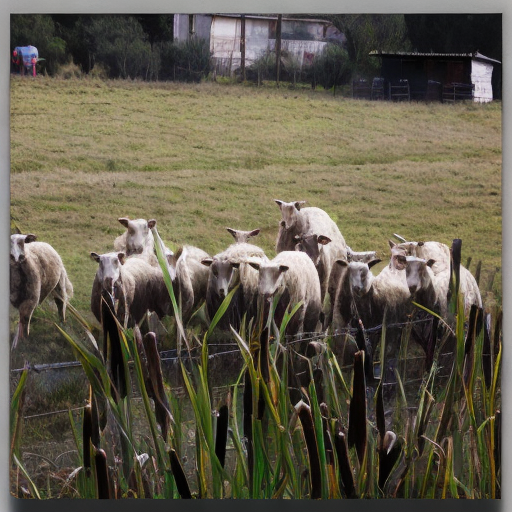}
        \end{subfigure}
        &
        \begin{subfigure}{0.245\textwidth}
            \centering
            \includegraphics[width=\linewidth]{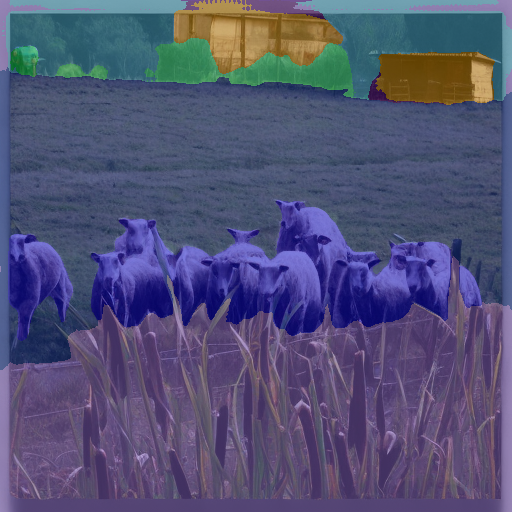}
        \end{subfigure}
        &
        \begin{subfigure}{0.245\textwidth}
            \centering
            \includegraphics[width=\linewidth]{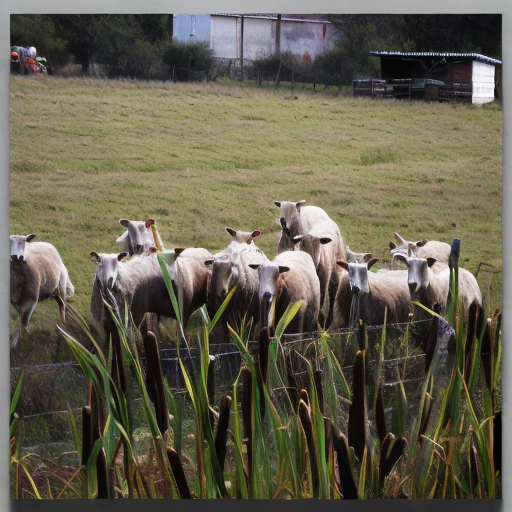}
        \end{subfigure}
        &
        \begin{subfigure}{0.245\textwidth}
            \centering
            \includegraphics[width=\linewidth]{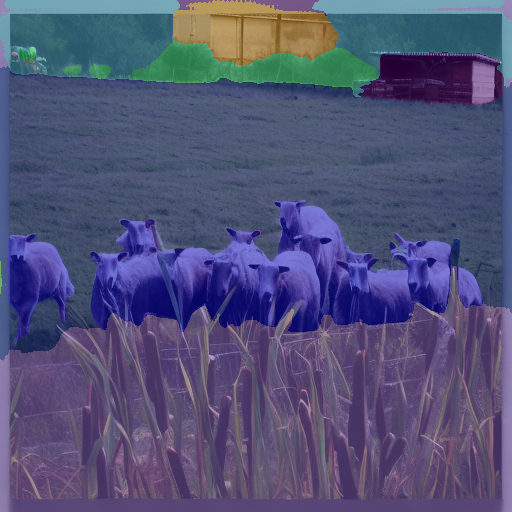}
        \end{subfigure} \\


    \end{tabular}

    \caption{Visual comparison of the semantic preservation of PerCo (SD) across various bit-rates on the MSCOCO-30k dataset (\texttt{000000442539}), using the ViT-Adapter segmentation network (Chen~\etal ICLR 2023). Global conditioning: "a herd of sheep standing in a field next to a fence".}
    \label{fig:vis_segmentation}
\end{figure*}

\textbf{Semantic preservation.} Finally, in~\cref{fig:vis_segmentation}, we visualize the semantic preservation capabilities of PerCo (SD) across all tested bit-rates.

\newpage

\end{document}